\documentclass[letterpaper, 10 pt, conference]{ieeeconf}  

\IEEEoverridecommandlockouts                              

\usepackage{graphicx}

\usepackage{amsmath} 
\usepackage{amssymb}  
\usepackage{tikz}
\usepackage{array}
\newcolumntype{C}[1]{>{\centering\arraybackslash}p{#1}}
\usepackage{comment}
\usepackage{color}
\usepackage{bbold}
\usepackage{booktabs}
\usepackage{lipsum}  
\usepackage{pifont}
\usepackage{bbm}

\usepackage{caption}
\usepackage{subcaption}

\usepackage{algpseudocode}
\usepackage{algorithm}
\usepackage{amsmath}
\usepackage[colorlinks]{hyperref}
\usepackage[capitalize]{cleveref}

\DeclareMathOperator*{\argmin}{\arg\!\min}

\title{\LARGE \bf
Discovering Multiple Algorithm Configurations
}

\newcommand\blfootnote[1]{%
  \begingroup
  \renewcommand\thefootnote{}\footnote{#1}%
  \addtocounter{footnote}{-1}%
  \endgroup
}
\author{Leonid Keselman$^{1}$ and Martial Hebert$^{1}$
\thanks{$^{1}$Leonid Keselman and Martial Hebert are with the Robotics Institute, School of Computer Science, Carnegie Mellon University, Pittsburgh PA, 15232 USA
        {\tt\small \{lkeselma,hebert\}@cs.cmu.edu}}
}

\begin{document}

\maketitle
\thispagestyle{empty}
\pagestyle{empty}

\begin{abstract}
Many practitioners in robotics regularly depend on classic, hand-designed algorithms. Often the performance of these algorithms is tuned across a dataset of annotated examples which represent typical deployment conditions. Automatic tuning of these settings is traditionally known as algorithm configuration. In this work, we extend algorithm configuration to automatically discover multiple modes in the tuning dataset. Unlike prior work, these configuration modes represent multiple dataset instances and are detected automatically during the course of optimization. We propose three methods for mode discovery: a post hoc method, a multi-stage method, and an online algorithm using a multi-armed bandit. Our results characterize these methods on synthetic test functions and in multiple robotics application domains: stereoscopic depth estimation, differentiable rendering, motion planning, and visual odometry. We show the clear benefits of detecting multiple modes in algorithm configuration space. 

\end{abstract}

\section{Introduction}\label{sec:intro}
\blfootnote{Further info at \url{https://leonidk.github.io/modecfg}}
Autonomous integrated systems often depend on a multitude of algorithms interacting with each other and their external environment. Despite the recent popularity of deep, end-to-end trained models~\cite{openai2019learning}, robotic systems often depend on hand-designed algorithms in several parts of the processing stack. They include motion planning~\cite{moveit2014}, algorithms involved in sensing~\cite{realsense2017} and  simultaneous location and mapping~\cite{dso2016}. As systems become more sophisticated, they often  accumulate more methods and with them, more parameters that need to be set and configured by the system designers. 

Often the developers of these methods can discover viable configurations by hand but leave many settings open to configuration by eventual users. Intuitively, these can include settings can control smoothing, performance and run-time. Ideal settings in noise-free environments can vary dramatically from those required in noisy settings. Likewise, different configurations may exist for optimal online and offline performance. Tuning such settings to work well in a deployment environment remains a challenge for many autonomous systems. Without proper tuning, components that are expected to be reliable may unexpectedly fail. 

\begin{figure}[th]
  \centering
  \includegraphics[width=\linewidth]{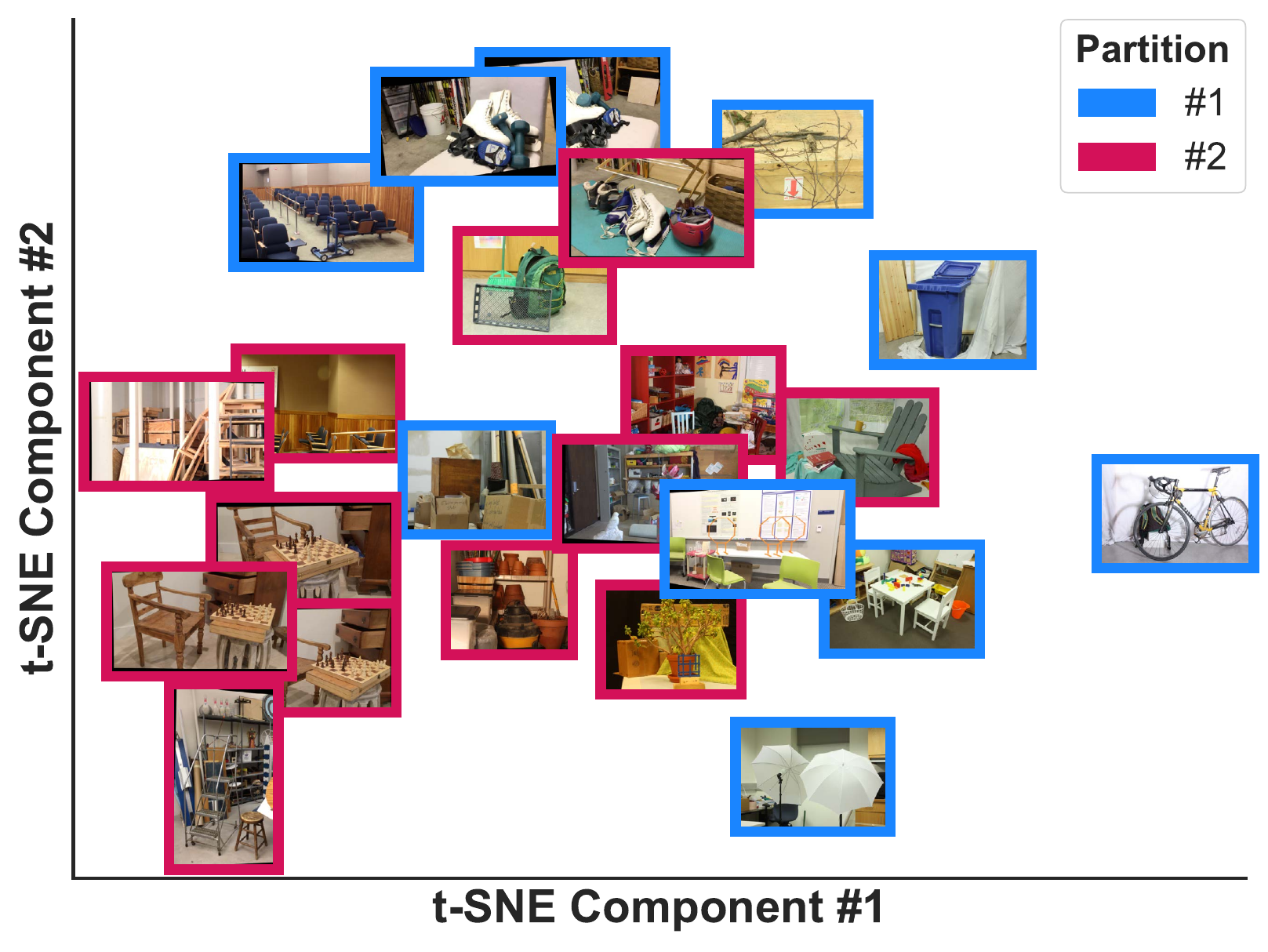}
  \caption{\textbf{Partitioned Configurations.} Instead of finding a single algorithm configuration for an entire dataset, we partition the dataset during the process of optimization and find a different configuration for each partition. }
  \label{fig:synth_viz}
\end{figure}

While it is possible to tune settings by hand, it is also possible to use automated methods to find potential configurations. In classic computer science literature, this was done to optimize runtime performance~\cite{Rice1976TheAS} and was known as the algorithm selection (or configuration) problem. Researchers have shown how automated tuning can quickly and robustly improve the performance of even common tools such as compilers~\cite{ansel:pact:2014}. In an era with many benchmarks~\cite{Geiger2012CVPR,Menze2015CVPR,10.1007/978-3-319-11752-2_3} and challenge competitions~\cite{krotkov2018darpa}, algorithm tuning is performed on a validation set made to approximate the performance of the final test set, and ensure the best possible performance for proposed techniques. To demonstrate consistency and generalizability, researchers often report performance under a single configuration.

Considering multiple configurations can greatly expand the applicability of a particular method~\cite{makatura2021paretogamuts}. In the case of multi-objective optimization, a Pareto set exists, where progress on any one objective regresses performance on another objective. As such, providing multiple configurations is often done by hardware vendors~\cite{realsense2017}, and selecting between multiple configurations in robotics is also well studied~\cite{Hu-RSS-17,pmlr-v78-hu17a,8593778}, along with selecting from ensembles of solvers~\cite{7139517} and motion-planners~\cite{7487136}. 

In this work, we propose to discover multiple viable configurations while an algorithm configuration is being automatically tuned over a dataset by some black-box optimizer~\cite{hansen2016cma,loshchilov2016cmaes}. By noticing correlations between data in response to new configurations, we detect multiple algorithm modes. Working in algorithm configuration space enables generalization across several problem domains, as they do not depend on domain specific features. This allows our method, with fixed settings, to show benefits in multiple application areas (see \cref{sec:apps}). We show how multiple modes can be found online (\cref{sec:online}) and how they can be used to guard against outliers (\cref{sec:stereo}). 

\section{Related Work}\label{sec:related}
There is a long history of work in algorithm configuration and related areas. Originally studied for performance optimization~\cite{Rice1976TheAS}, algorithm configuration has been well studied in the SAT solver community~\cite{10.5555/2832249.2832351}.  These extensions include large evaluations~\cite{doi:10.1080/10556788.2020.1808977}, taxonomies of methods~\cite{https://doi.org/10.48550/arxiv.2202.01651} and methods which adapt the algorithm configuration over a series of time-steps~\cite{https://doi.org/10.48550/arxiv.2105.08541,10.1145/3449726.3459578,https://doi.org/10.48550/arxiv.2205.13881}, assuming the algorithms used have different temporal properties. 

Since robotics applications and datasets are reasonably expensive operations (compared to purely synthetic tasks), our work is related to work on extremely few function evaluations, typically on the order of 100~\cite{https://doi.org/10.48550/arxiv.2103.10321}. Typically this is studied in the Machine Learning community as hyperparameter search, finding configurations for ideal neural network configuration and training ~\cite{Kotthoff2019,loshchilov2016cmaes}. 

Our work is most closely related to portfolio-based algorithm configuration literature~\cite{10.5555/1630659.1630927,10.5555/3000001.3000101,10.1145/1538902.1538906}. These methods often design a different configuration for each instance of the problem in their dataset (e.g. each data example)~\cite{5365884,10.5555/2898607.2898641,10.5555/1860967.1861114}. Similar to our work, they cluster methods into groups. However, their clusters are derived from domain-specific feature spaces, while we use the response of the instances to new configurations. Our use of domain-specific features is limited to test time deployment, using supervised training to target our obtained algorithm configuration clusters. 

In the Machine Learning community, the problem of coreset discovery~\cite{https://doi.org/10.48550/arxiv.1906.01827} is related to our approach. Coreset discovery finds representative examples from a dataset to focus training time on a subset of the data. Most related, methods exist for online discovery of such sets~\cite{yoon2022online}. Of note, our contribution is orthogonal to coreset research, as our method could benefit from coreset methods, which would serve to give us a smaller subset of data to evaluate at each iteration. Specifically, coreset methods for clustering~\cite{https://doi.org/10.48550/arxiv.1702.08248} could enable more function evaluation steps in a fixed budget of time and give better resulting minima. 

In computer vision, some recent work has explored estimating algorithm configurations for classic algorithms on a local level, operating on patches within an image~\cite{https://doi.org/10.48550/arxiv.2112.09318}.

\section{Method}\label{sec:method}
Our approach consists of evaluating algorithm configurations from a black box optimizer (\cref{sec:blackbox}) across a dataset of examples for the given algorithm (\cref{alg:optimizer}). Building upon this baseline, we propose three methods of partitioning the data during optimization: Post hoc (\cref{alg:posthoc}), Staged (\cref{alg:staged}), and Online (\cref{alg:online}). 

For our experiments, we always use two partitions, even when there are more known modes (\cref{sec:snyth}). This avoids exploring the area of instance-specific algorithm configuration and minimizes our risk of overfitting to the dataset and overstating our performance. We typically report results between the initialization (the known defaults for an algorithm) and the oracle. Our oracle is defined as awarding the best known configuration for each individual datum across all optimization runs.

\subsection{Partitioning}\label{sec:partition}

The optimal partition for a given number of partitions $K$, with $M$ algorithm configurations over $N$ datapoints can be formulated via 0-1 Integer Linear Programming where $c_{i,j}$ corresponds to the quality of datum $i$ with configuration $j$.  

\begin{gather*}
\min\quad c_{i,j} \, x_{i,j}  \\
\begin{aligned}
\textup{subject to}\qquad\\
x_{i,j}  &\in  \{0,1\}\\
             \sum_{j=1}^M x_{i,j}  &=  1 \\ 
             \sum_{j=1}^M \mathbb{1}{\left[ (\sum_{i=1}^N x_{i,j}) > 0 \right] }   &\leq  K \\ 
\end{aligned}
\end{gather*}

One can also exhaustively evaluate all ${M \choose K}$ partitions. Our experiments do exhaustive evaluation for $K=2$ (as with all results in this paper) and use the optimization formulation with larger numbers of partitions. We solve the optimization problem with a recent solver~\cite{Huangfu2018} and implement the indicator variables using the BigM modeling trick. 

As an alternative, we evaluate using a clustering method such as k-Means~\cite{10.1145/1772690.1772862} on a normalized matrix $\tilde{X}$, where each row has zero mean and unit variance. Cluster centers are in the space of the history of evaluated configurations and each row is a datum's response to the history of evaluated configurations. Clustering approaches treat the algorithm configuration history as a feature and group results which behave similarly, but may not be optimally partitioned.

\begin{figure}[thb]
  \centering
  \includegraphics[width=\linewidth]{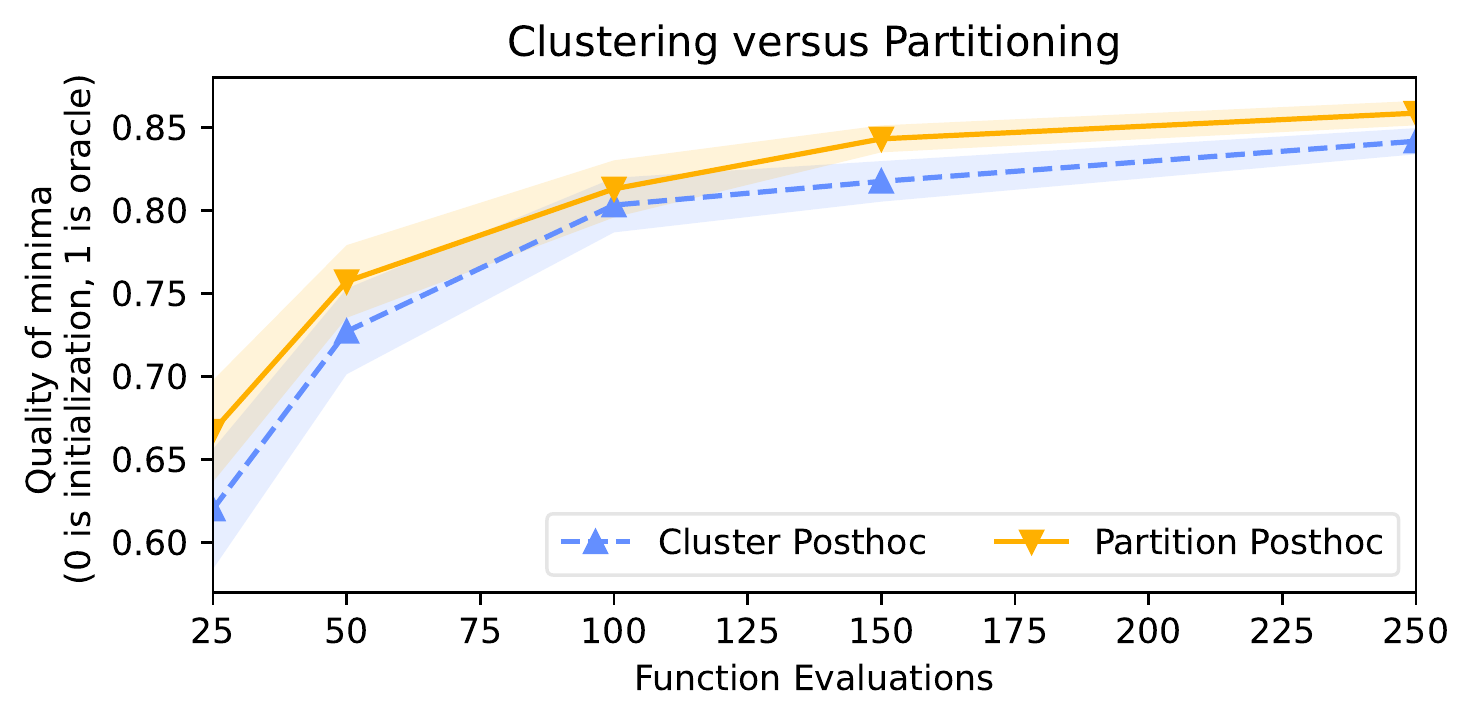}
  \caption{\textbf{Partitioning vs K-Means Clustering} on the stereoscopic depth experiments described in \cref{sec:stereo}.}
  \label{fig:part_v_clust}
\end{figure}

\subsection{Black Box Optimizer}\label{sec:blackbox}
Our method is generic to the choice of black box optimizer, also known as gradient-free optimization. For example, one could use random search, which is known to be a strong baseline in higher dimensional optimization~\cite{46180}. On the other extreme, if one has expensive function evaluations, one could fit a surrogate function to the data and optimize its expected minima, as is done in Bayesian Optimization \cite{https://doi.org/10.48550/arxiv.1807.02811}. Effective black box optimizers in practice often combine a plethora of optimizers~\cite{ansel:pact:2014,nevergrad}.

We use CMA-ES, an evolutionary method with a multivariate Gaussian model~\cite{hansen2016cma}. CMA-ES is known for algorithm configuration search in robotics~\cite{doi:10.1126/science.aal5054} and is widely used in other algorithm configuration comparisons~\cite{https://doi.org/10.48550/arxiv.2105.08541}. CMA-ES is convenient in only requiring an initial configuration and a $\sigma$ in parameter space. When tuning existing algorithms, reasonable prior configurations are often available. Approaches which focus on modeling a bounded volume ~\cite{https://doi.org/10.48550/arxiv.1807.02811,nevergrad}) can be wasteful. All of our parameter search is for non-negative parameters, so we transform our search space with $\log(x)$ to perform unconstrained optimization, making our search operate on order-of-magnitude scale for all parameters.

\subsection{Post hoc Partitioning}\label{sec:posthoc}
The post hoc method is simple and straightforward: perform black box on the dataset as a whole, noting each datum's response to each configuration. Afterwards, partition the data following \cref{sec:partition}. This approach allows clearest evaluation against the non-partitioned CMA-ES baseline, which it outperforms in all of our experiments. \Cref{alg:posthoc} outlines the ~\emph{Post hoc} method in detail. As a method with no change in exploration, it can be be run on existing single mode optimization or simply using coherently evaluated random configurations~\cite{46180}, as in \cref{sec:realsense}. 

\subsection{Staged Partitioning}\label{sec:staged}
In staged partitioning, we spend half of the function evaluations exploring the space to find adequate minimia, and we spend half of the function evaluations exploiting the discovered partitions in isolation. While the scale of a particular problem may suggest different balance of exploration and exploitation stages, we use a ratio of $\frac{1}{2}$ for our experiments. \Cref{alg:staged} performs partitioning in the middle of optimization and tunes the results for each partition. This enables more explicit exploitation of the partitions, at the expense of less exploration time to find good partitions. 

\subsection{Online Partitioning}\label{sec:online}
To balance exploration and exploitation in an online fashion, one could use a multi-armed bandit. \Cref{alg:online} dynamically assigns data points to partitions during the course of optimization. Since CMA-ES only evaluates relative order, we can readily switch data assigned to each partition during the course of evaluation. For the online method, we setup a multi-armed bandit (MAB) for each datum. Since our function evaluations are unbounded, we use classic Thompson Sampling~\cite{10.2307/2332286,https://doi.org/10.48550/arxiv.1707.02038} with a Gaussian distribution. We perform one CMA-ES step to sample the space, and use that to initialize the distributions for each arm of the bandits identically. Each iteration, we sample a partition assignment for each bandit. That datum then evaluates the configuration given by that optimizer and records its result for that arm. The optimizers record the mean cost of the data assigned to them that iteration. 

This approach allows us to simultaneously perform multiple optimizations and partition assignments on the fly.

\begin{algorithm}
\caption{\strut Optimize algorithm configuration over a dataset}\label{alg:optimizer}
\begin{algorithmic}[1]
\Require {$x_0$} Initial Configuration
\Require {$f_{1...N}(x)$} dataset queries for algorithm
\Require {$M$} Maximum number of function evaluations

\Procedure{Optimize}{$x_0,f_{1...N}(x),M$}
\For{$i \gets 1$ to $N$}
    \State $x_i\gets\Call{Optimizer Candidate}$

    \For{$j \gets 1$ to $N$} \Comment{Evaluate all data}
    \State {$X_{i,j}$ $\gets$ {$f_{j}(x_i)$}}
    \EndFor
    \State {$g_i$ $\gets$ {$\Call{Mean}{X_i}$}} 
    \State $\Call{Optimizer Tell}{g_i}$\Comment{Report average}
\EndFor
\State {$Y_i$ $\gets$ $\Call{Mean}{X_{i,j}}  $} \Comment{Per configuration scores}
\State {$x$ $\gets$ {$x_{\argmin(Y_i)}$}} \Comment{Best configuration}
\EndProcedure
\end{algorithmic}
\end{algorithm}

\begin{algorithm}
\caption{\strut Finding modes with post hoc partitioning}\label{alg:posthoc}
\begin{algorithmic}[1]
\Require {$K$} Number of partitions
\Ensure {$x_{1..K}$} Per partition configurations
\Ensure {$c_{1..N}$} Per datum partition assignments

\Procedure{Posthoc}{$x_0,f_{1...N}(x),M,K$}
\State {$\Call{Optimize}{x_0,f_{1...N}(x),M}$} 
\State {$c_j$ $\gets$ {$\Call{Partition}{X^T,K}$}} \Comment{Get partitions for data}

\State {$Y_k$ $\gets$ $\Call{Mean}{X_{i,(c_j=k)}}  $} \Comment{Per partition scores}
\State {$x_k$ $\gets$ $x_{\argmin(Y_k)}$} \Comment{Configuration for partitions}

\EndProcedure
\end{algorithmic}
\end{algorithm}

\begin{algorithm}
\caption{\strut Finding modes with staged partitioning}\label{alg:staged}
\begin{algorithmic}[1]
\Procedure{Staged}{$x_0,f_{1...N},M,K$}
\State {$\Call{Post hoc}{x_0,f_{1...N}(x),\frac{M}{2},K}$} 

\For{$k \gets 1$ to $K$} \Comment{Separate optimization}
\State {$\Call{Optimize}{x_0,f_{c_j = k}(x),\frac{M}{2}}$} 
\EndFor
\EndProcedure
\end{algorithmic}
\end{algorithm}

\begin{algorithm}
\caption{\strut Finding modes with online partitioning}\label{alg:online}
\begin{algorithmic}[1]
\Procedure{Online}{$x_0,f_{1...N},M,K$}
\State {$B_k \gets \Call{Bandit}{N}$} \Comment{$K$ arms for each datum}
\State {$OPT_k \gets \Call{Optimizer()}{}$} \Comment{$K$ optimizers}
\For{$i \gets 1$ to $M$}
    \For{$j \gets 1$ to $N$}
    \State {$b_j$ $\gets$ {$\Call{Bandit Pull}{B_j}$}} \Comment{Sample bandit}
    \EndFor
\For{$m \gets 1$ to $K$} \Comment{Separate Evaluation}
    \State $x_{i,k} \gets OPT_k \Call{Candidate}$
    \State $y_{i,k} \gets \Call{Mean}{f_{b_j = k}(x_{i,k})}$
    \State $ OPT_k \Call{Tell}{y_{i,k}}$
\EndFor
\EndFor
\State {$c_j \gets \Call{Best Arm}{ B_j}$}
\State {$y_k \gets \Call{Best Config}{OPT_k}$}
\EndProcedure

\end{algorithmic}
\end{algorithm}


\section{Experimental Results}\label{sec:apps}
We evaluate our approach on several application domains. We start with a synthetic function whose structure and modes are known and is quick to evaluate. This enables us to characterize our different methods of finding partitions. We then proceed to show successful benefits to robotics methods like stereoscopic depth generation~\cite{realsense2017}, differentiable rendering ~\cite{keselman2022fuzzy}, motion planning~\cite{Lavalle98rapidly-exploringrandom}, and visual odometry~\cite{dso2016}.

\subsection{Synthetic Function}\label{sec:snyth}
A synthetic function allows us to characterize our methods across arbitrary many dimensions and modes. Our synthetic function has $K$ modes, each the sum of $N$ hard-to-optimize functions, leading to a simulation of $K N$ data points. We use four hard-to-optimize functons: Ackley, Griewank, Rastrigin, Zakharov (for details of these functions and their visualizations see \cite{Lv2018}), and rescale them to have a minima of value zero, a random rotation, and to have a maximum value of around one near the minima. This paradigm and these functions can be generated in arbitrary many dimensions, allowing us to understand how these partitioning methods scale as algorithm hyper-parameters scale from two to forty dimensions. 

For the synthetic function optimization, the staged method works best across most dimensions and number of function evaluations. Close behind, especially with fewer evaluations, is the post hoc method. In contrast, our online bandit method is typically only slightly better than the single mode baseline. Of note is that all methods begin to perform better with hundreds of function evaluations, suggesting that the improved performance of the partitioning may come from improved efficiency in low numbers of evaluations, and not the multi-modal nature of the synthetic function. 

  \begin{figure}[thpb]
  \centering
  \begin{subfigure}[t]{0.85\linewidth}
  \centering
  \includegraphics[width=\textwidth]{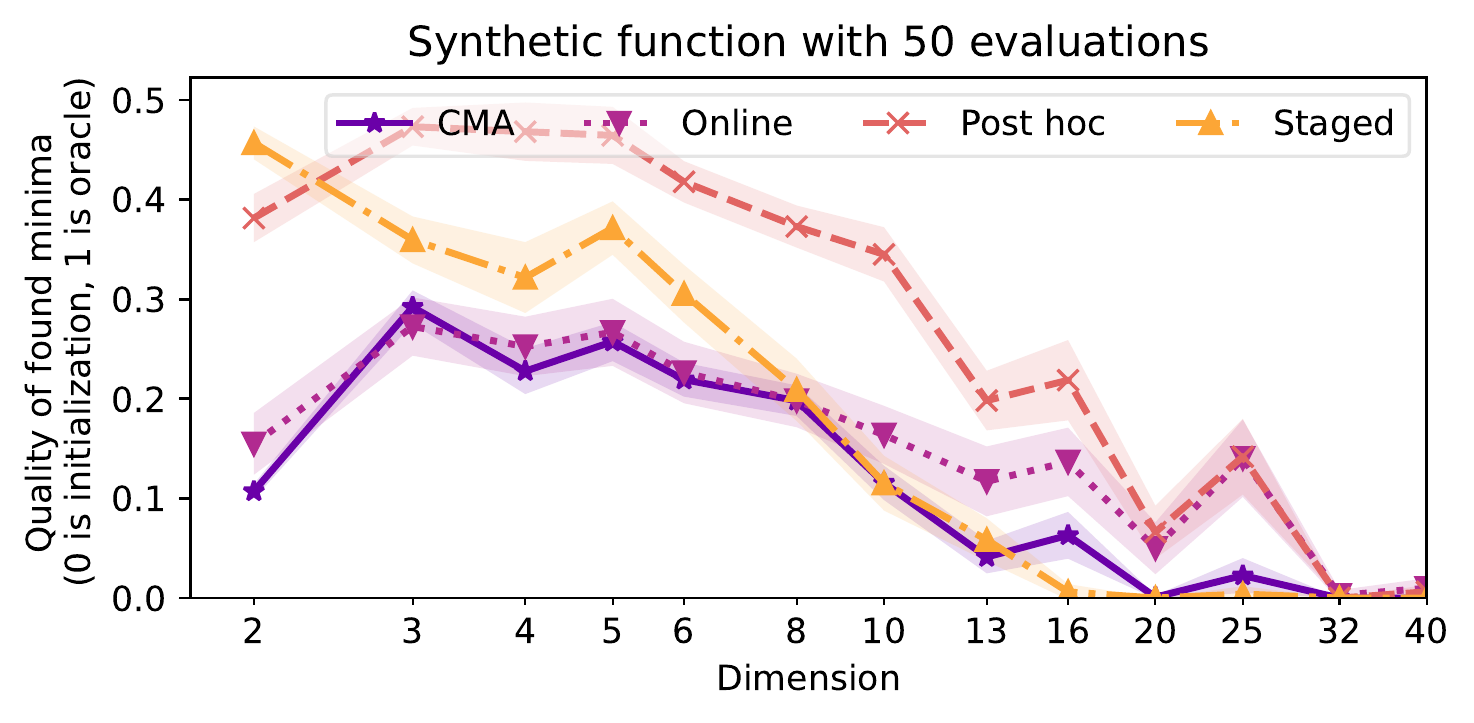}
  \end{subfigure}
  \hfill
  \begin{subfigure}[t]{0.85\linewidth} 
    \centering
    \includegraphics[width=\textwidth]{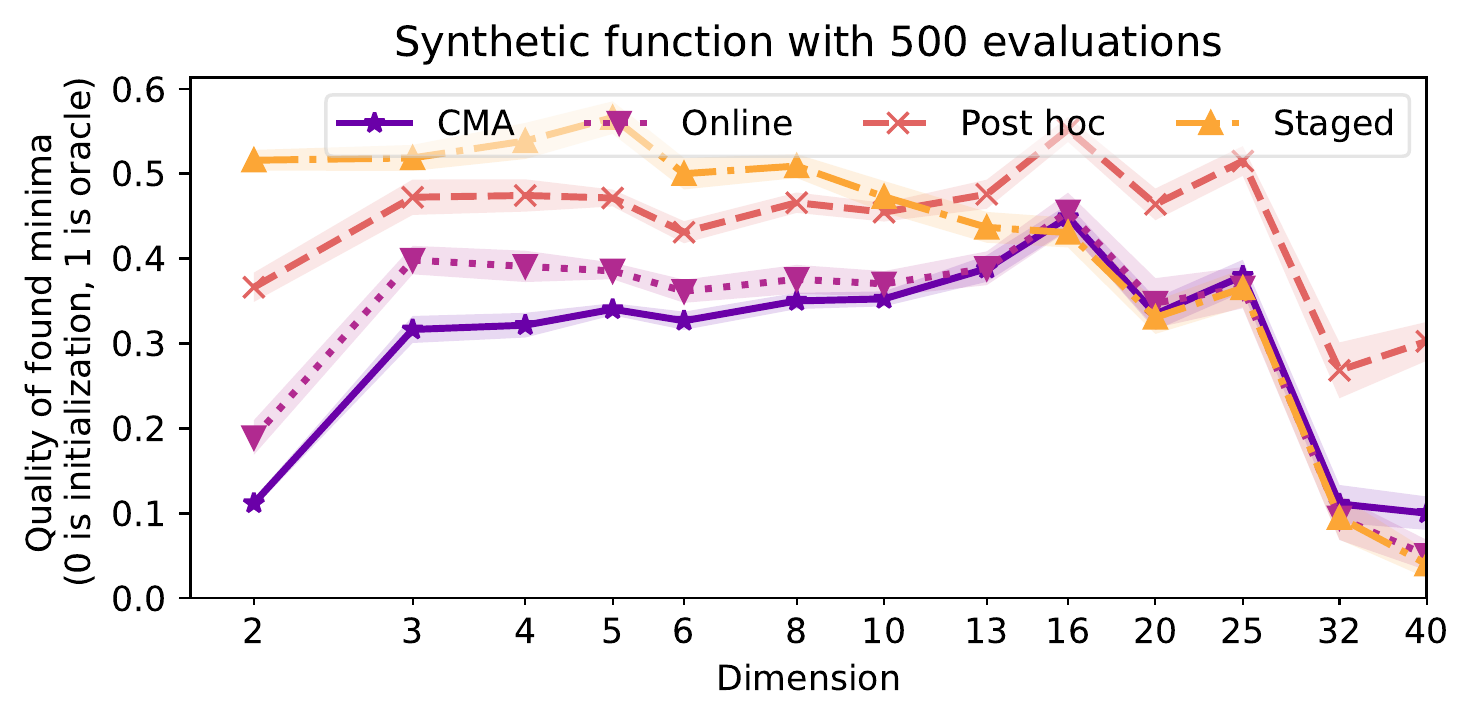}
  \end{subfigure}
  \caption{\textbf{Synthetic Function Partitioning} (\cref{sec:snyth}). Graphs shows the quality of the best found minima for all methods, between the initial configuration and an oracle.  Shaded regions indicate standard error of the mean. }
  \label{fig:synth_conv}
\end{figure}

\subsection{Dense Stereo Matching}\label{sec:stereo}
\begin{figure}[tpb]
  \centering
  \includegraphics[width=\linewidth]{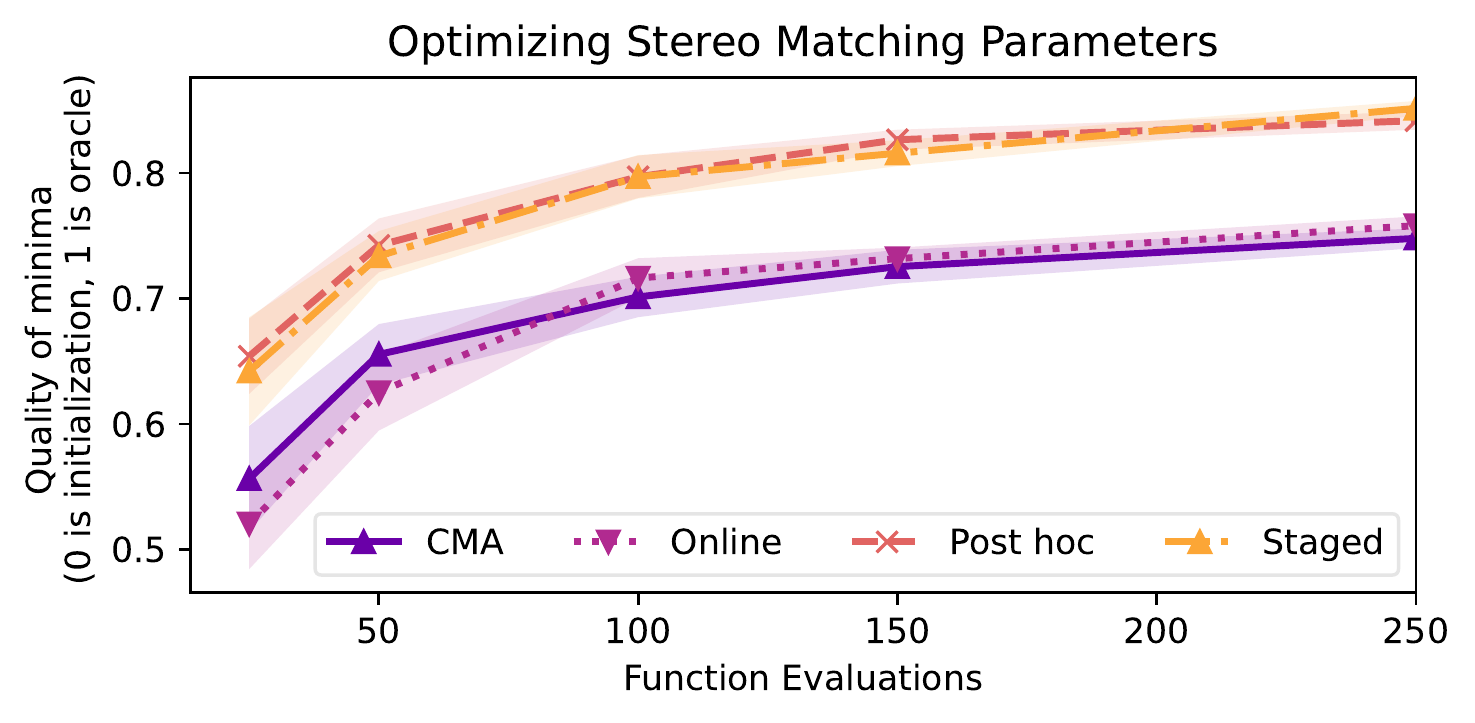}
  \caption{ \textbf{Dense Stereo Matching Partitioning} quality on the training set. The posthoc and staged methods perform well while the online method is indistinguishable from  CMA-ES. }
  \label{fig:opt_stereo}
\end{figure}

\begin{figure}[thpb]
  \centering
  \begin{subfigure}[t]{\linewidth} 
    \centering
    \includegraphics[width=0.27\textwidth]{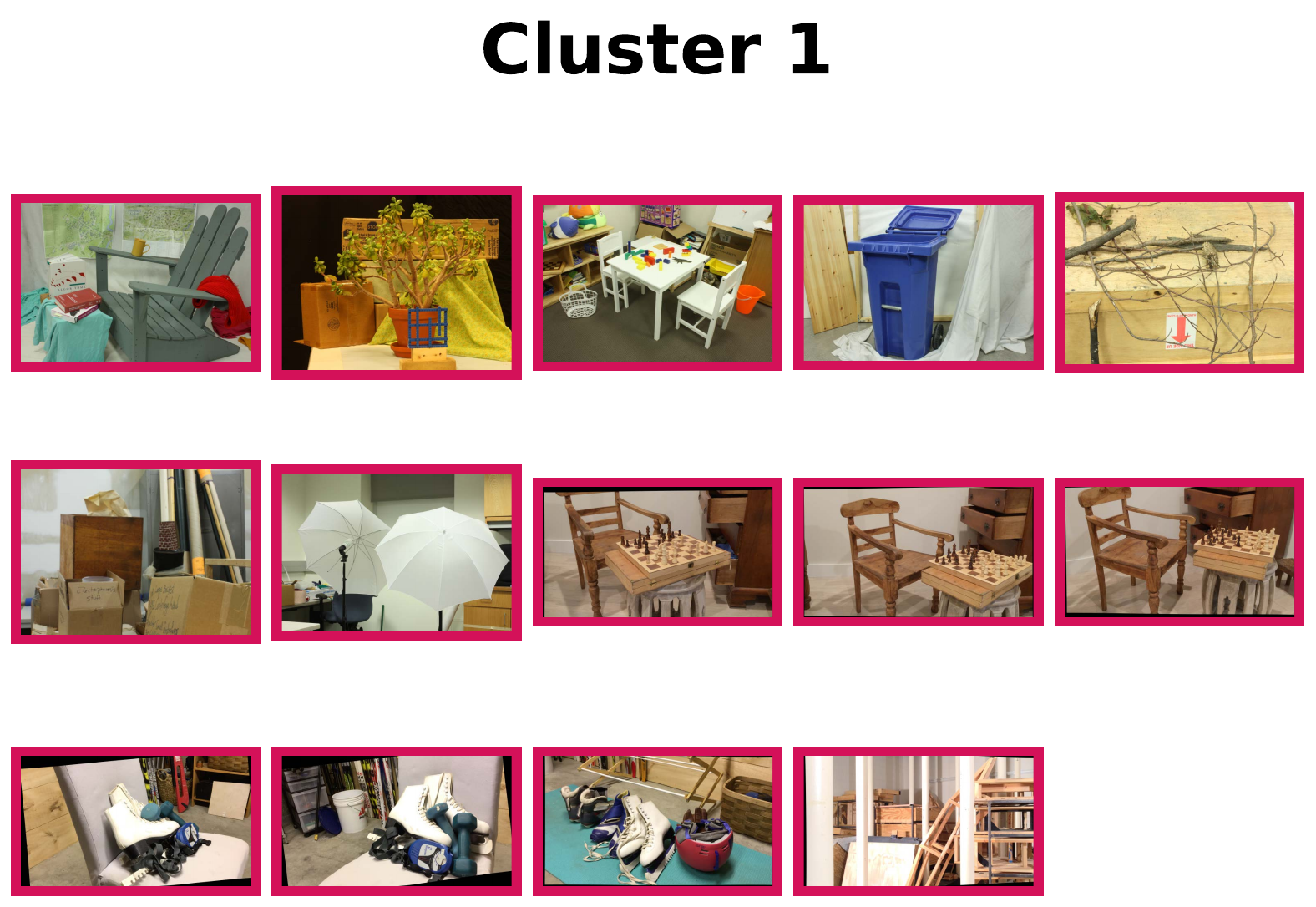}
    \includegraphics[width=0.27\textwidth]{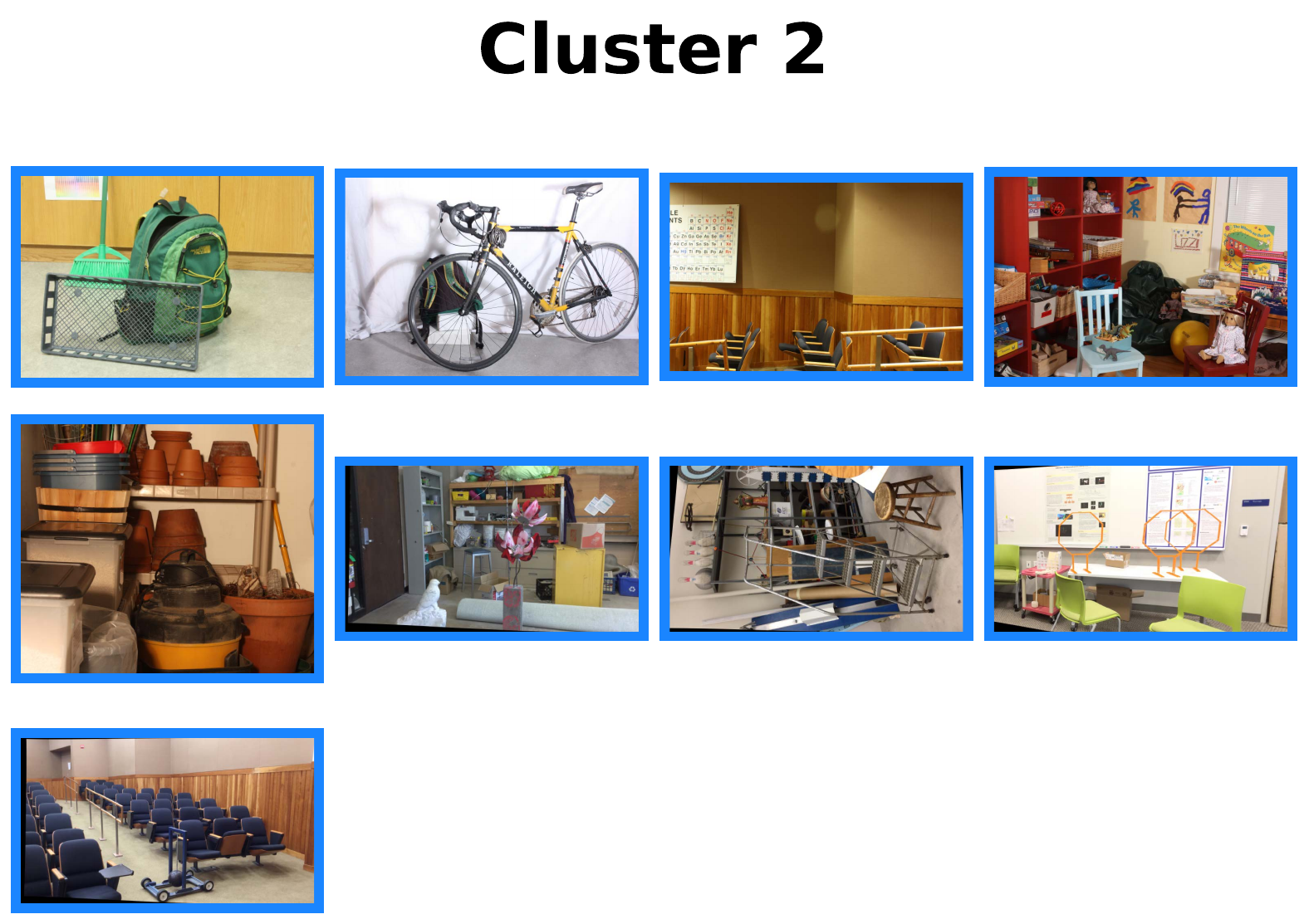}
    \includegraphics[width=0.42\textwidth]{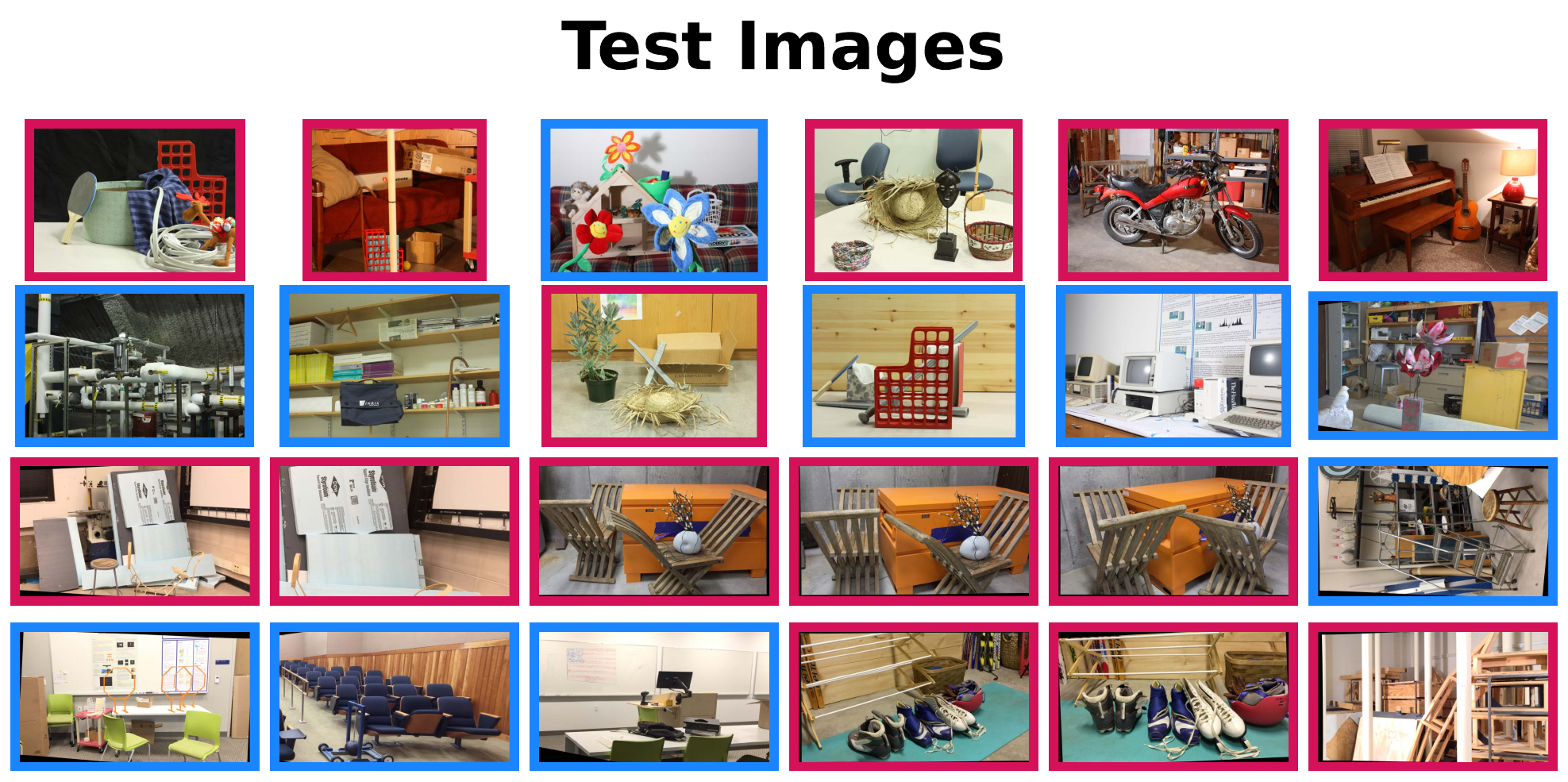}
    \label{fig:stereo_split}
    \caption{\textbf{Middlebury Stereo Data} with training data shown as partitioned by our staged method after 250 evaluations. The test set is shown with the predicted classification colors for each image. }
  \end{subfigure}
  \begin{subfigure}[t]{\linewidth} 
    \includegraphics[width=\linewidth]{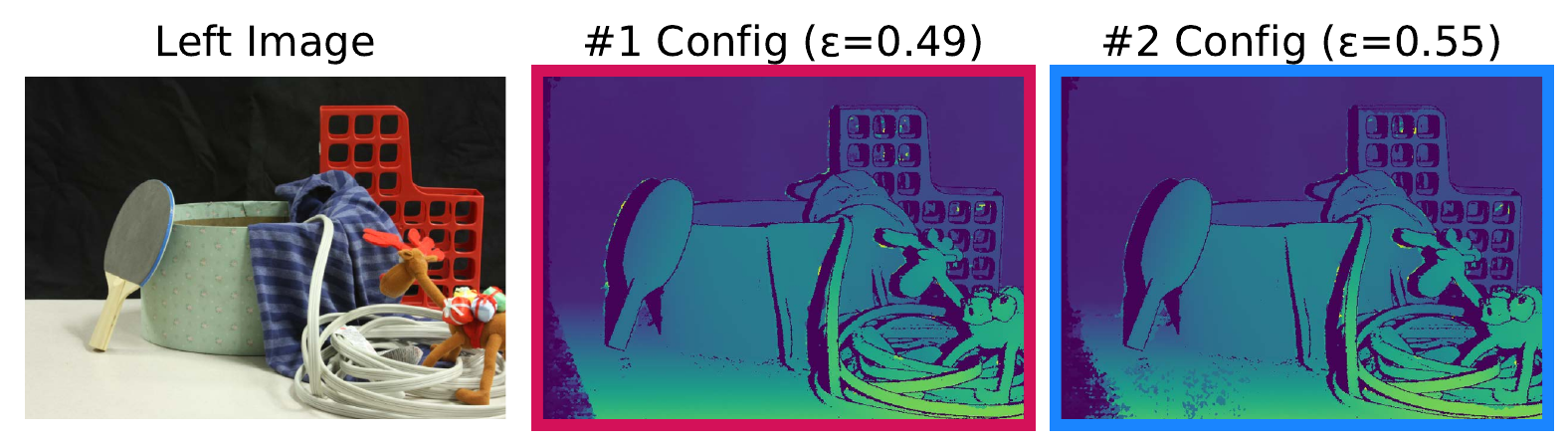}
  \includegraphics[width=\linewidth]{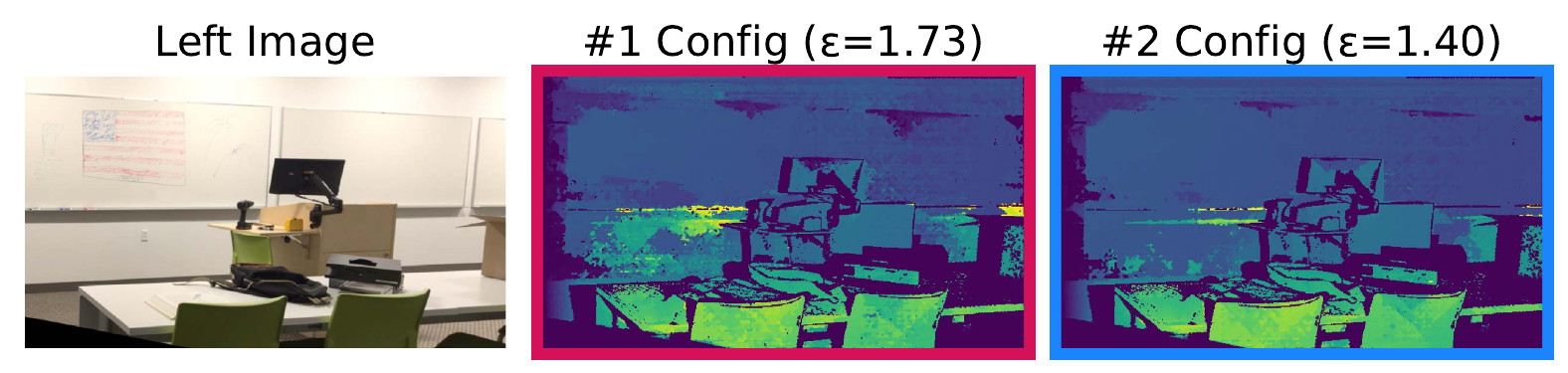}
    \caption{\textbf{Qualitative examples of test set results} with our classifier correctly assigning the better configuration to each example. Disparity error shown in parenthesis for each configuration and example.}
    \label{fig:stereo_vis}
    \end{subfigure}
  \begin{subfigure}[t]{\linewidth} 
    \includegraphics[width=\linewidth]{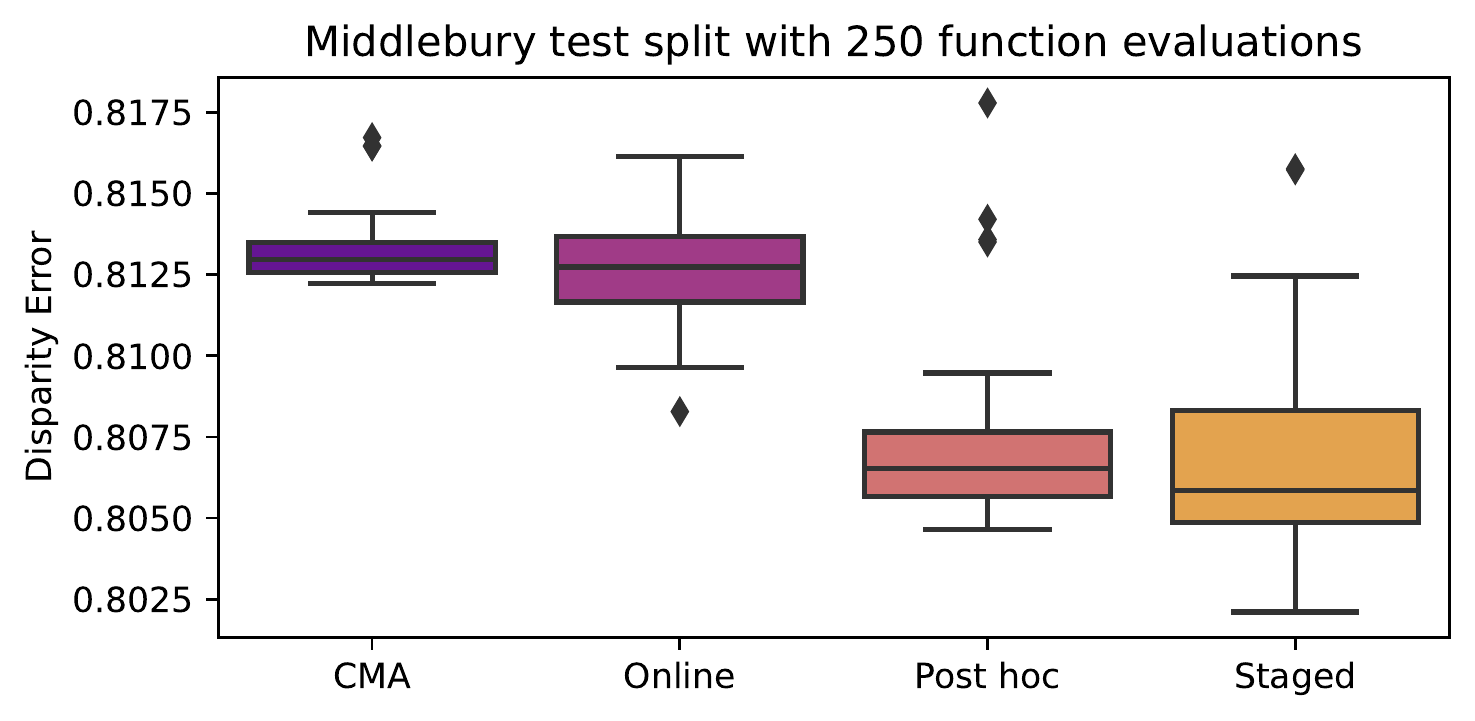}
    \caption{\textbf{Test set performance} based on the configurations preferred by our supervised classifier described in \cref{sec:stereo}.}
    \label{fig:stereo_test_box}
\end{subfigure}
  \caption{\textbf{Dense Stereo Matching Test Set Performance}. }
  \label{fig:stereo_out3}
\end{figure}

Robotics applications often use stereoscopic depth sensors. Here we optimize the performance a classic Dense Stereo Matching method, namely Semi-Global Block Matching (SGBM)~\cite{4359315} as implemented by OpenCV~\cite{opencv_library}. We obtain 47 image pairs by combining the Middlebury 2014 and 2021 Stereo datasets~\cite{10.1007/978-3-319-11752-2_3}. We split the data into 23 training examples and 24 test examples, shown in \cref{fig:stereo_split}. The algorithm settings control the regularization of the SGBM algorithm, the post-processing filters used to cleanup the data, and the block size used for initial matching. Results of the four methods on the training set are shown in \cref{fig:opt_stereo}. 

In deploying the discovered configurations to new data, we show the efficacy of a simple supervised classifier.The classifier used is $k$-nearest neighbors with a $k=1$, returning the partition index to be used. We use a pre-trained neural network's top level feature space as the feature space. Specifically we use SqueezeNet 1.1~\cite{https://doi.org/10.48550/arxiv.1602.07360} pre-trained on ImageNet in PyTorch~\cite{NEURIPS2019_9015} and its 512 dimensional space for images.

The test set performance is improved with partitioning, as shown in \cref{fig:stereo_test_box} with quantitative estimates and \cref{fig:stereo_vis} with two qualitative examples from the test set. 

We find that the optimal hyperparameter configuration typically focuses on regularization and filtering. The first configuration usually has less regularization, but a more aggressive filter to discard bad matches, while the second configuration has more regularization and less aggressive filters to discard bad data. 


\subsection{Differentiable Rendering}\label{sec:fuzzy}

We optimize the parameters of a recent differentiable renderer~\cite{keselman2022fuzzy}. Our dataset includes 20 sequences from the KITTI odometry dataset~\cite{Geiger2012CVPR} and 20 synthetic shapes. The KITTI sequences use k-Means to build a quick model of 10 consecutive LIDAR frames, from the center of the scene looking out. In contrast, the synthetic sequences all have the object densely in front camera. The differentiable render has four hyperparameters, two controlling the sharpness of the silhouettes, one controlling surface smoothness and one controlling how opaque objects are. We optimize all four for depth and silhouette accuracy, similar to the original paper. 

\begin{figure}[tpb]
  \centering
  \begin{subfigure}[t]{\linewidth}
  \centering
  \includegraphics[width=\linewidth]{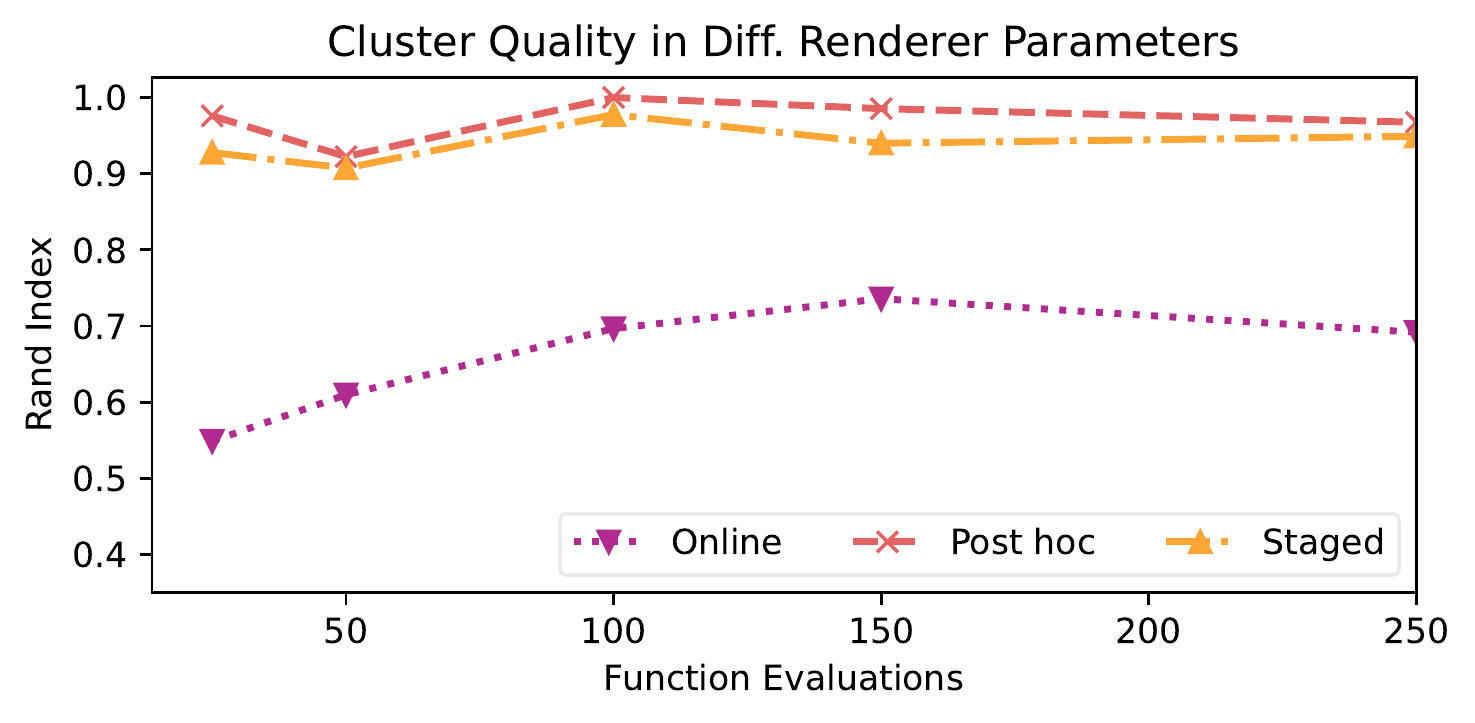}
  \caption{\textbf{Partitioning Accuracy} in splitting the KITTI data from  synthetic data in \cref{sec:fuzzy}. All methods show some success. }
  \label{fig:fm_partition_qual}
  \end{subfigure}
  \begin{subfigure}[t]{\linewidth} 
  \centering
  \includegraphics[width=\linewidth]{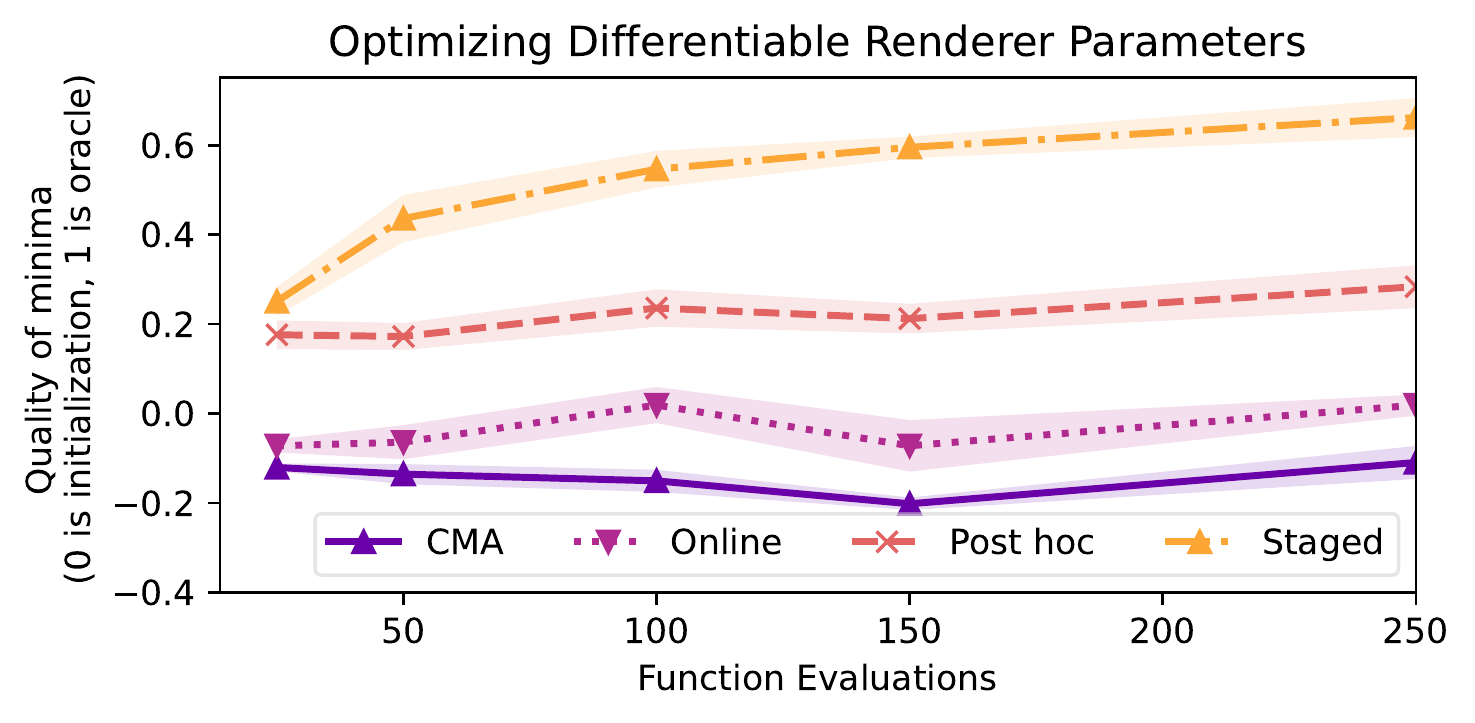}
    \caption{\textbf{Differentiable Renderer Partitioning} hyper-parameter optimization for depth and silhouette fidelity. \Cref{sec:fuzzy}.  In this case, reasonable defaults restrict the progress of the baseline while the exploitation-focused staged optimization performs best. }
  \end{subfigure}
  \caption{\textbf{Differentiable Renderer Experiments}}
  \label{fig:opt_fuzz}
\end{figure}

In rendering, the optimizer is unable to find a better single mode configuration than the initialization. However, all proposed methods show a statistically significant improvement over the baseline, with the staged method performing the best. In addition, we report the ability of the methods to properly partition the disparate datasets in \cref{fig:fm_partition_qual}. 


\subsection{Motion Planning}\label{sec:planning}
\begin{figure}[thpb]
  \centering
  \includegraphics[width=\linewidth]{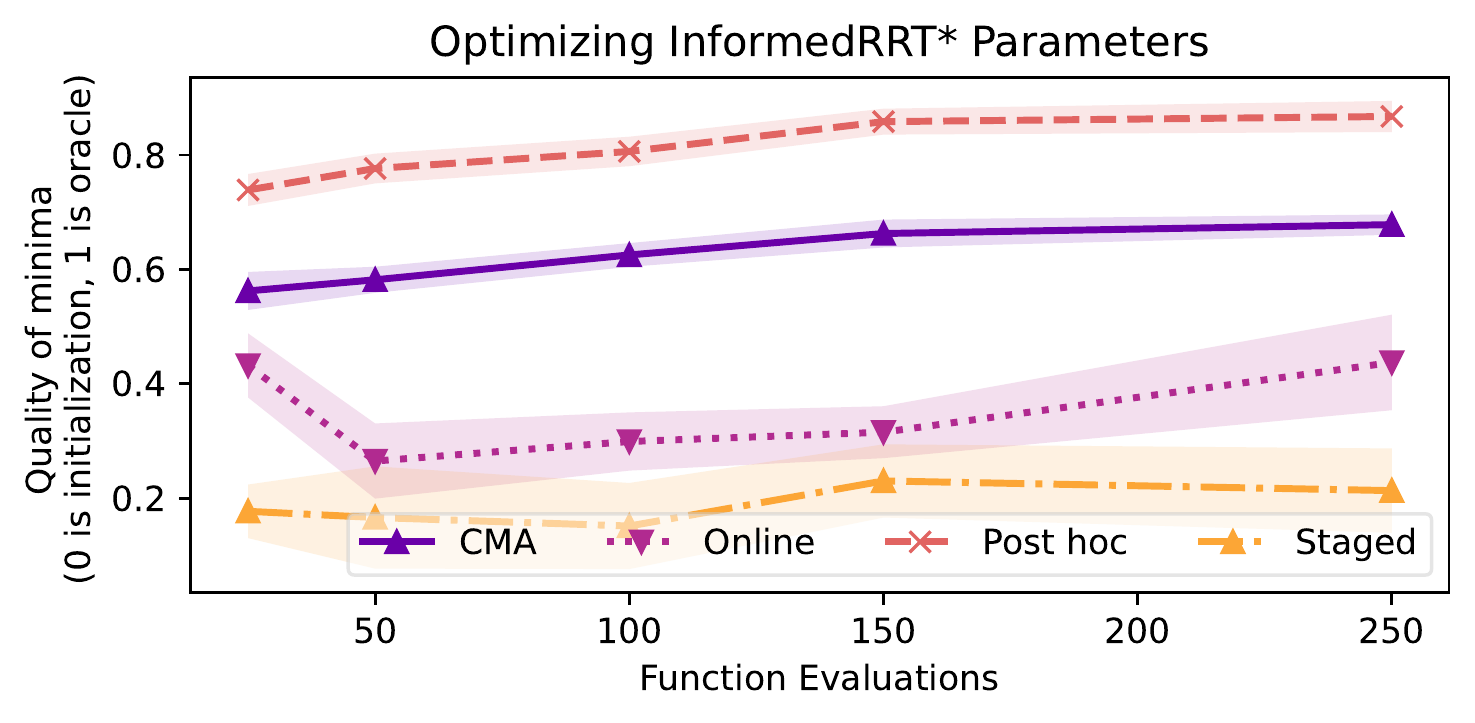}
  \caption{\textbf{Motion Planner Partitioning} on a set of environments and planar planning tasks. \cref{sec:planning} for details.  }
  \label{fig:opt_plan}
\end{figure}

We evaluate a popular motion planning method, Informed RRT*~\cite{6942976} in the Sampling-Based Motion Planner Testing Environment~\cite{lai2021SbpEnv}. We setup three start-goal pairs for three testing environments.  We use the geometric mean of runtime (as estimated by the number of expanded notes) and performance (as estimated by the quality of the first found solution). This balance of runtime and quality is essential to obtaining an interested configuration. Results are shown in \cref{fig:opt_plan}, with only the post hoc method outperforming the baseline. Other methods perform poorly, and we suspect the problem is insufficient samples and lack of exploration in the online and staged methods; especially as RRT*-based planners are stochastic, making evaluations noisy. 

We find that the optimal partition finds different parameters focused on the goal sampling frequency (0.2 and 0.3; single mode 0.26) and the rewiring radius (1500 vs 7500; single mode 6000). Of our three start/goal pairs in each of three different environments, optimal partitioning typically grouped one environment together.

We also performed some experiments with RRdT*~\cite{lai_rrdt2019}, which we report briefly report. Often, one partition would focus on a configuration that frequently spawned new trees, while the other focused on expanding existing trees.

As it is unclear how to parameterize motion planning goals and environments for supervised classification, we were unable to do  experiments on a hold-out test set. 

\subsection{Visual Odometry}\label{sec:slam}
We perform  experiments on a subset of the TUM VI Visual-Inertial Dataset~\cite{schubert2018vidataset} using DM-VIO~\cite{stumberg22dmvio}. DM-VIO has many parameters but we focus on five (points, immature points, min frames, max frames, max optimization steps). TUM VI has 5 environments, and we select the third sequence from each environment as our dataset. 

\begin{figure}[tpb]
  \centering
  \begin{subfigure}[t]{\linewidth}
    \centering
    \includegraphics[width=\linewidth]{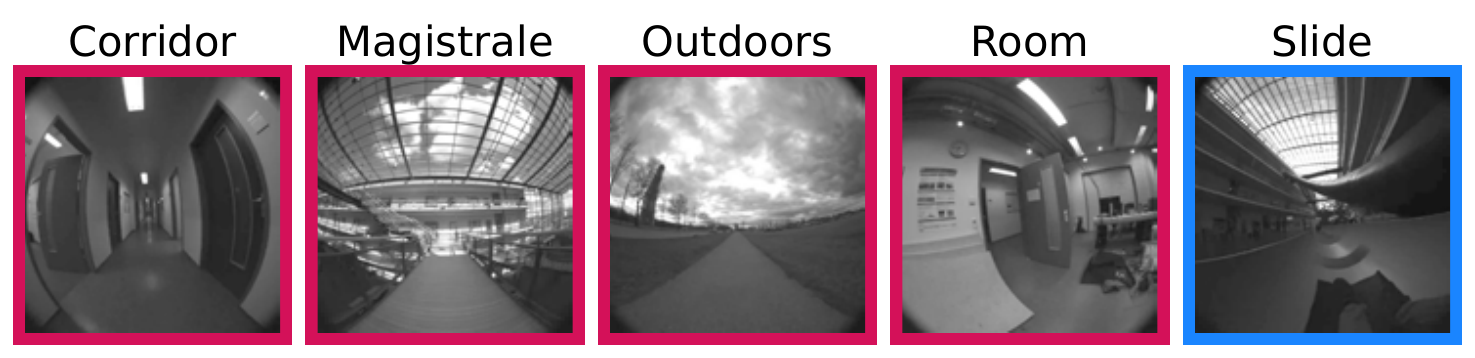}
    \caption{Partitioned TUM-VI Visual Odometry Dataset  }
    \label{fig:slam_vis}
  \end{subfigure}
  
  \begin{subfigure}[t]{\linewidth} 
    \centering
    \includegraphics[width=\linewidth]{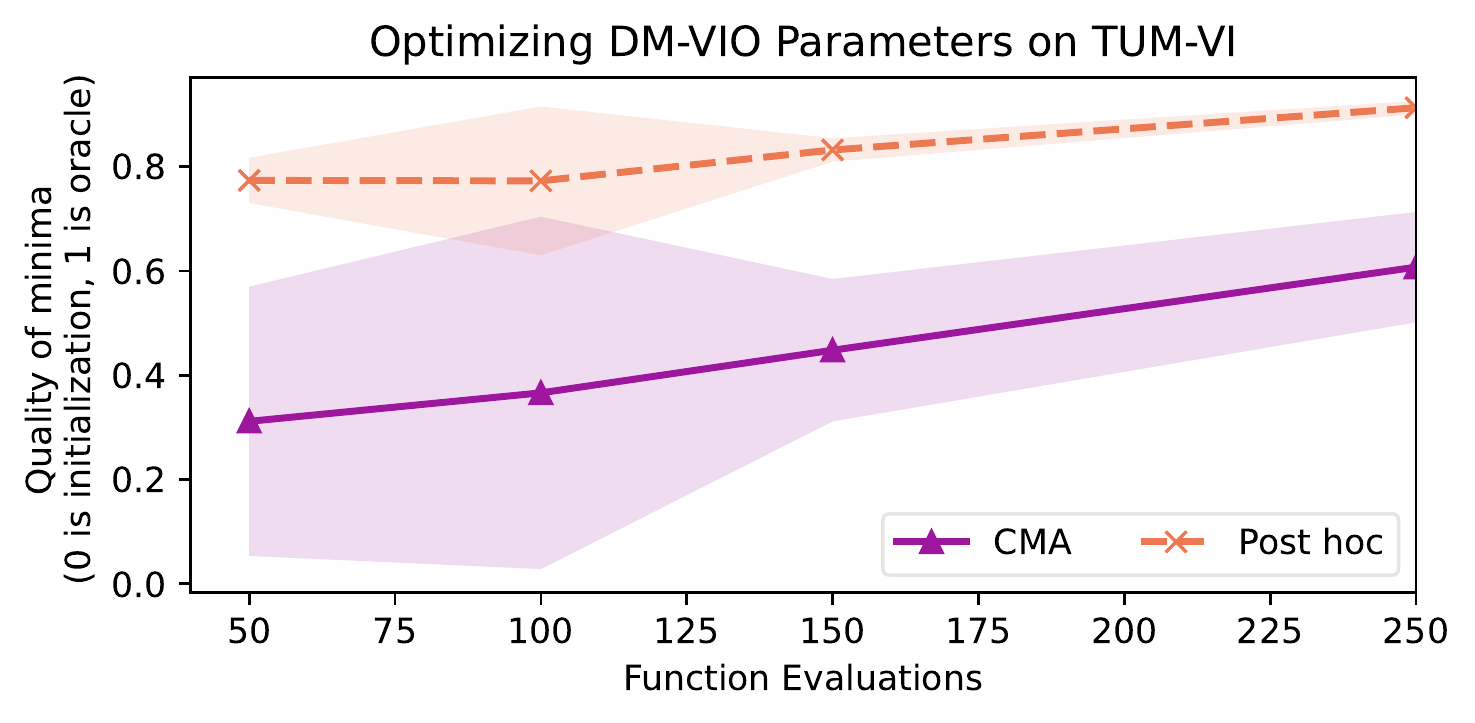}
    \caption{Post-hoc clustering improves DM-VIO performance. }
    \label{tab:slam_table}
  \end{subfigure}
  \caption{\textbf{Visual-Inertial Odometry Experiments}}
  \label{fig:opt_vio}
\end{figure}

We prioritize a geometric mean of runtime (frame time) and quality (best-aligned absolute pose error~\cite{grupp2017evo}), while  penalizing trajectories which do not complete successfully. Results are shown in \cref{fig:opt_vio}. Reliably, the algorithm partitions a separate configuration for all but the \textit{slide3} sequence, which includes fast motion through a closed pipe. The slides partition is the best single mode, while the alternative partition uses fewer points (100 vs 350) and fewer frames (2-4 vs 3-5) as it does not need to handle the difficult high-velocity, highly occluded sequence. 

As our VO partitions depend on properties of the sequence, we were unable to construct a reasonable test set based on the first frame. Instead, our multiple configurations may be used in on-the-fly configuration selection~\cite{pmlr-v78-hu17a},

\subsection{Commercial Depth Sensor}\label{sec:realsense}
\begin{figure}[tpb]
  \centering
  \includegraphics[width=\linewidth]{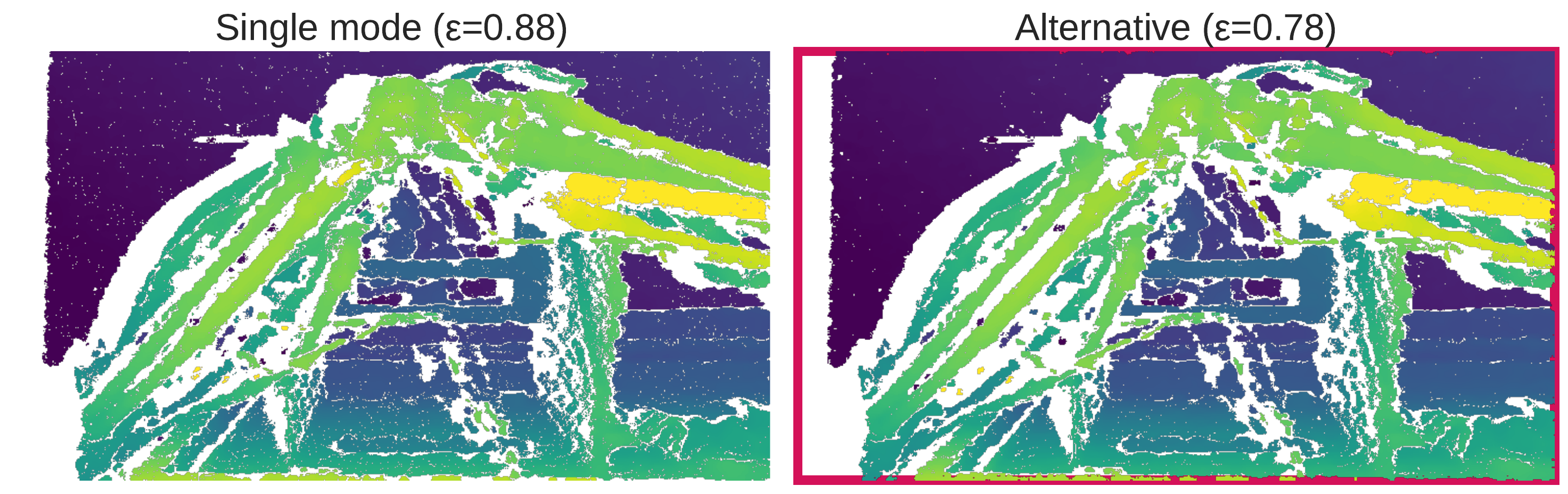}
    \includegraphics[width=\linewidth]{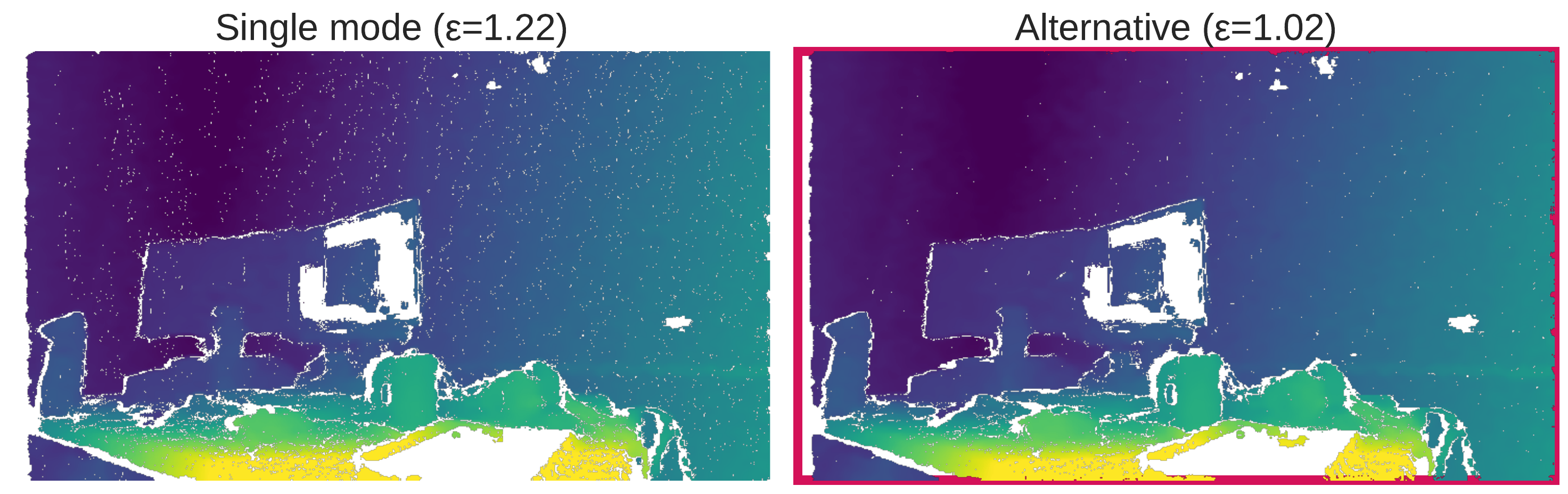}
  \includegraphics[width=\linewidth]{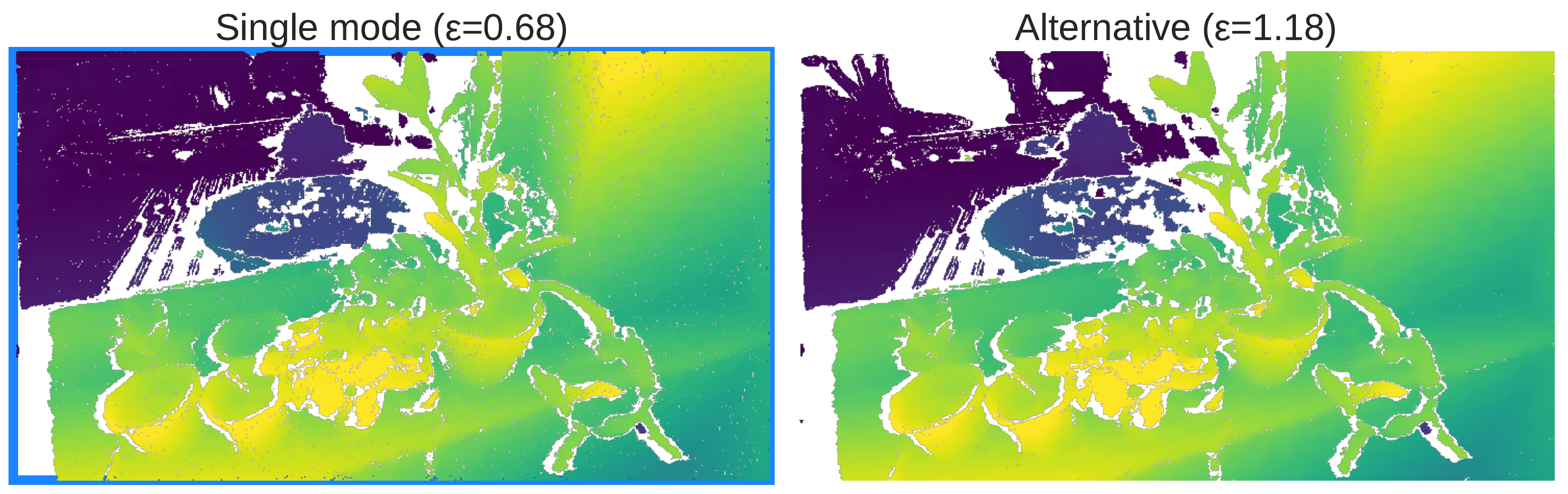}
    \includegraphics[width=\linewidth]{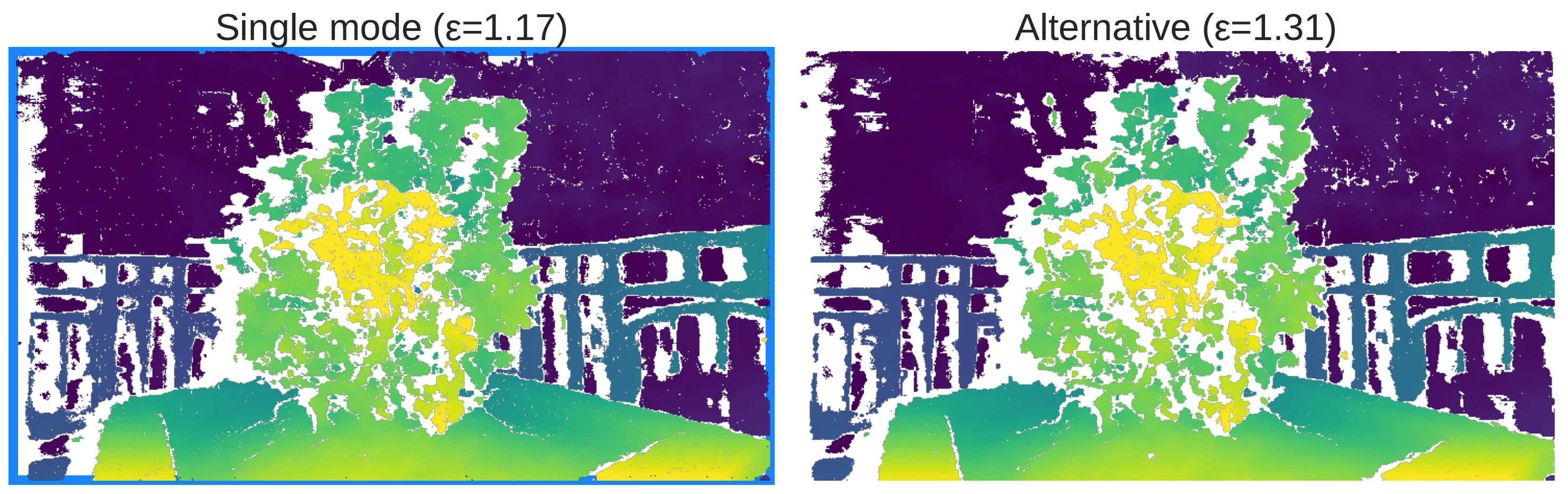}
  \caption{\textbf{Intel RealSense D435 Partitioning} with 500 randomly generated configurations on 10 scenes. Two are shown, with both configurations (and its error). \cref{sec:realsense}. }
  \label{fig:realsense}
\end{figure}

Lastly, we demonstrate our partitioning method on a Intel RealSense D435~\cite{realsense2017} and its 35 parameters for estimating depth. We generate a set of 500 randomly  configurations. We evaluate all configurations on 10 scenes, for which we have collected their pseudo ground truths using a moving laser pattern~\cite{realsense2017}. We partition the configurations using the post hoc method. Results for four scenes are shown in \cref{fig:realsense}. 

The optimal $K=2$ partition included the best single mode configuration as well, allowing us to show it and the alternative configuration. The single mode configuration produced small holes, but dense results outdoors. While the alternative configuration produced smoother, denser walls in indoor environments in exchange for more artifacts outdoors.

\section{Discussion}\label{sec:disc}
Many algorithms in robotics operate in environments with multiple modes. These natural partitions are easy to understand, and can be discovered naturally by analyzing how different datums respond to different algorithm configurations. The modes were found because they affected algorithm response, not because they happened to be grouped together in some domain-specific feature space. 

All the proposed methods for partitioning show some efficacy. Across the board, the post hoc method works well. This is likely due our extremely small number of evaluations for algorithm configuration~\cite{https://doi.org/10.48550/arxiv.2103.10321}, leading to more benefits for exploration. The online method typically performs poorly in this setting and it is possible that more sophisticated bandit algorithms~\cite{menard:hal-01475078} could perform better.

Our experiments focused on two partitions for all methods. Even when problems had more modes by construction, two partitions were able to clearly improve performance. 

\section{Conclusion}\label{sec:conc}
Automatically finding modes during the course of algorithm configuration is a viable way to improve algorithm performance in several different areas of robotics. More study is needed to understand what typical algorithm modes exist and how such modal configurations might be used long-term deployed autonomous systems. 


\bibliographystyle{IEEEtran}
\bibliography{IEEEabrv,mybibfile}

\begin{thebibliography}{10}
\providecommand{\url}[1]{#1}
\csname url@rmstyle\endcsname
\providecommand{\newblock}{\relax}
\providecommand{\bibinfo}[2]{#2}
\providecommand\BIBentrySTDinterwordspacing{\spaceskip=0pt\relax}
\providecommand\BIBentryALTinterwordstretchfactor{4}
\providecommand\BIBentryALTinterwordspacing{\spaceskip=\fontdimen2\font plus
\BIBentryALTinterwordstretchfactor\fontdimen3\font minus
  \fontdimen4\font\relax}
\providecommand\BIBforeignlanguage[2]{{%
\expandafter\ifx\csname l@#1\endcsname\relax
\typeout{** WARNING: IEEEtran.bst: No hyphenation pattern has been}%
\typeout{** loaded for the language `#1'. Using the pattern for}%
\typeout{** the default language instead.}%
\else
\language=\csname l@#1\endcsname
\fi
#2}}

\bibitem{openai2019learning}
OpenAI, M.~Andrychowicz, B.~Baker, M.~Chociej, R.~Jozefowicz, B.~McGrew,
  J.~Pachocki, A.~Petron, M.~Plappert, G.~Powell, A.~Ray, J.~Schneider,
  S.~Sidor, J.~Tobin, P.~Welinder, L.~Weng, and W.~Zaremba, ``Learning
  dexterous in-hand manipulation,'' 2019.

\bibitem{moveit2014}
\BIBentryALTinterwordspacing
D.~Coleman, I.~Sucan, S.~Chitta, and N.~Correll, ``Reducing the barrier to
  entry of complex robotic software: a moveit! case study,'' 2014. [Online].
  Available: \url{https://arxiv.org/abs/1404.3785}
\BIBentrySTDinterwordspacing

\bibitem{realsense2017}
\BIBentryALTinterwordspacing
L.~Keselman, J.~I. Woodfill, A.~Grunnet-Jepsen, and A.~Bhowmik, ``Intel
  realsense stereoscopic depth cameras,'' 2017. [Online]. Available:
  \url{https://arxiv.org/abs/1705.05548}
\BIBentrySTDinterwordspacing

\bibitem{dso2016}
\BIBentryALTinterwordspacing
J.~Engel, V.~Koltun, and D.~Cremers, ``Direct sparse odometry,'' 2016.
  [Online]. Available: \url{https://arxiv.org/abs/1607.02565}
\BIBentrySTDinterwordspacing

\bibitem{Rice1976TheAS}
J.~R. Rice, ``The algorithm selection problem,'' \emph{Adv. Comput.}, vol.~15,
  pp. 65--118, 1976.

\bibitem{ansel:pact:2014}
\BIBentryALTinterwordspacing
J.~Ansel, S.~Kamil, K.~Veeramachaneni, J.~Ragan-Kelley, J.~Bosboom, U.-M.
  O'Reilly, and S.~Amarasinghe, ``Opentuner: An extensible framework for
  program autotuning,'' in \emph{International Conference on Parallel
  Architectures and Compilation Techniques}, Edmonton, Canada, Aug 2014.
  [Online]. Available:
  \url{http://groups.csail.mit.edu/commit/papers/2014/ansel-pact14-opentuner.pdf}
\BIBentrySTDinterwordspacing

\bibitem{Geiger2012CVPR}
A.~Geiger, P.~Lenz, and R.~Urtasun, ``Are we ready for autonomous driving? the
  kitti vision benchmark suite,'' in \emph{Conference on Computer Vision and
  Pattern Recognition (CVPR)}, 2012.

\bibitem{Menze2015CVPR}
M.~Menze and A.~Geiger, ``Object scene flow for autonomous vehicles,'' in
  \emph{Conference on Computer Vision and Pattern Recognition (CVPR)}, 2015.

\bibitem{10.1007/978-3-319-11752-2_3}
D.~Scharstein, H.~Hirschm{\"u}ller, Y.~Kitajima, G.~Krathwohl,
  N.~Ne{\v{s}}i{\'{c}}, X.~Wang, and P.~Westling, ``High-resolution stereo
  datasets with subpixel-accurate ground truth,'' in \emph{Pattern
  Recognition}, X.~Jiang, J.~Hornegger, and R.~Koch, Eds.\hskip 1em plus 0.5em
  minus 0.4em\relax Cham: Springer International Publishing, 2014, pp. 31--42.

\bibitem{krotkov2018darpa}
E.~Krotkov, D.~Hackett, L.~Jackel, M.~Perschbacher, J.~Pippine, J.~Strauss,
  G.~Pratt, and C.~Orlowski, ``The darpa robotics challenge finals: Results and
  perspectives,'' in \emph{The DARPA Robotics Challenge Finals: Humanoid Robots
  To The Rescue}.\hskip 1em plus 0.5em minus 0.4em\relax Springer, 2018, pp.
  1--26.

\bibitem{makatura2021paretogamuts}
\BIBentryALTinterwordspacing
L.~Makatura, M.~Guo, A.~Schulz, J.~Solomon, and W.~Matusik, ``Pareto gamuts:
  Exploring optimal designs across varying contexts,'' \emph{ACM Transactions
  on Graphics (SIGGRAPH)}, vol.~40, no.~4, pp. 1--17, 8 2021. [Online].
  Available: \url{https://doi.org/10.1145/3450626.3459750}
\BIBentrySTDinterwordspacing

\bibitem{Hu-RSS-17}
H.~Hu and G.~Kantor, ``Introspective evaluation of perception performance for
  parameter tuning without ground truth,'' in \emph{Proceedings of Robotics:
  Science and Systems}, Cambridge, Massachusetts, July 2017.

\bibitem{pmlr-v78-hu17a}
\BIBentryALTinterwordspacing
------, ``Efficient automatic perception system parameter tuning on site
  without expert supervision,'' in \emph{Proceedings of the 1st Annual
  Conference on Robot Learning}, ser. Proceedings of Machine Learning Research,
  S.~Levine, V.~Vanhoucke, and K.~Goldberg, Eds., vol.~78.\hskip 1em plus 0.5em
  minus 0.4em\relax PMLR, 13--15 Nov 2017, pp. 57--66. [Online]. Available:
  \url{https://proceedings.mlr.press/v78/hu17a.html}
\BIBentrySTDinterwordspacing

\bibitem{8593778}
------, ``Compensating for context by learning local models of perception
  performance,'' in \emph{2018 IEEE/RSJ International Conference on Intelligent
  Robots and Systems (IROS)}, 2018, pp. 4629--4634.

\bibitem{7139517}
S.~Choudhury, S.~Arora, and S.~Scherer, ``The planner ensemble: Motion planning
  by executing diverse algorithms,'' in \emph{2015 IEEE International
  Conference on Robotics and Automation (ICRA)}, 2015, pp. 2389--2395.

\bibitem{7487136}
A.~Tallavajhula, S.~Choudhury, S.~Scherer, and A.~Kelly, ``List prediction
  applied to motion planning,'' in \emph{2016 IEEE International Conference on
  Robotics and Automation (ICRA)}, 2016, pp. 213--220.

\bibitem{hansen2016cma}
N.~Hansen, ``The cma evolution strategy: A tutorial,'' 2016.

\bibitem{loshchilov2016cmaes}
I.~Loshchilov and F.~Hutter, ``Cma-es for hyperparameter optimization of deep
  neural networks,'' 2016.

\bibitem{10.5555/2832249.2832351}
C.~Ans\'{o}tegui, Y.~Malitsky, H.~Samulowitz, M.~Sellmann, and K.~Tierney,
  ``Model-based genetic algorithms for algorithm configuration,'' in
  \emph{Proceedings of the 24th International Conference on Artificial
  Intelligence}, ser. IJCAI'15.\hskip 1em plus 0.5em minus 0.4em\relax AAAI
  Press, 2015, p. 733–739.

\bibitem{doi:10.1080/10556788.2020.1808977}
\BIBentryALTinterwordspacing
N.~Hansen, A.~Auger, R.~Ros, O.~Mersmann, T.~Tušar, and D.~Brockhoff, ``Coco:
  a platform for comparing continuous optimizers in a black-box setting,''
  \emph{Optimization Methods and Software}, vol.~36, no.~1, pp. 114--144, 2021.
  [Online]. Available: \url{https://doi.org/10.1080/10556788.2020.1808977}
\BIBentrySTDinterwordspacing

\bibitem{https://doi.org/10.48550/arxiv.2202.01651}
\BIBentryALTinterwordspacing
E.~Schede, J.~Brandt, A.~Tornede, M.~Wever, V.~Bengs, E.~Hüllermeier, and
  K.~Tierney, ``A survey of methods for automated algorithm configuration,''
  2022. [Online]. Available: \url{https://arxiv.org/abs/2202.01651}
\BIBentrySTDinterwordspacing

\bibitem{https://doi.org/10.48550/arxiv.2105.08541}
\BIBentryALTinterwordspacing
T.~Eimer, A.~Biedenkapp, M.~Reimer, S.~Adriaensen, F.~Hutter, and M.~Lindauer,
  ``Dacbench: A benchmark library for dynamic algorithm configuration,'' 2021.
  [Online]. Available: \url{https://arxiv.org/abs/2105.08541}
\BIBentrySTDinterwordspacing

\bibitem{10.1145/3449726.3459578}
\BIBentryALTinterwordspacing
F.~Ye, C.~Doerr, and T.~B\"{a}ck, ``Leveraging benchmarking data for informed
  one-shot dynamic algorithm selection,'' in \emph{Proceedings of the Genetic
  and Evolutionary Computation Conference Companion}, ser. GECCO '21.\hskip 1em
  plus 0.5em minus 0.4em\relax New York, NY, USA: Association for Computing
  Machinery, 2021, p. 245–246. [Online]. Available:
  \url{https://doi.org/10.1145/3449726.3459578}
\BIBentrySTDinterwordspacing

\bibitem{https://doi.org/10.48550/arxiv.2205.13881}
\BIBentryALTinterwordspacing
S.~Adriaensen, A.~Biedenkapp, G.~Shala, N.~Awad, T.~Eimer, M.~Lindauer, and
  F.~Hutter, ``Automated dynamic algorithm configuration,'' 2022. [Online].
  Available: \url{https://arxiv.org/abs/2205.13881}
\BIBentrySTDinterwordspacing

\bibitem{https://doi.org/10.48550/arxiv.2103.10321}
\BIBentryALTinterwordspacing
C.~Ansotegui, M.~Sellmann, T.~Shah, and K.~Tierney, ``Learning how to optimize
  black-box functions with extreme limits on the number of function
  evaluations,'' 2021. [Online]. Available:
  \url{https://arxiv.org/abs/2103.10321}
\BIBentrySTDinterwordspacing

\bibitem{Kotthoff2019}
\BIBentryALTinterwordspacing
L.~Kotthoff, C.~Thornton, H.~H. Hoos, F.~Hutter, and K.~Leyton-Brown,
  \emph{Auto-WEKA: Automatic Model Selection and Hyperparameter Optimization in
  WEKA}.\hskip 1em plus 0.5em minus 0.4em\relax Cham: Springer International
  Publishing, 2019, pp. 81--95. [Online]. Available:
  \url{https://doi.org/10.1007/978-3-030-05318-5_4}
\BIBentrySTDinterwordspacing

\bibitem{10.5555/1630659.1630927}
K.~Leyton-Brown, E.~Nudelman, G.~Andrew, J.~McFadden, and Y.~Shoham, ``A
  portfolio approach to algorithm select,'' in \emph{Proceedings of the 18th
  International Joint Conference on Artificial Intelligence}, ser.
  IJCAI'03.\hskip 1em plus 0.5em minus 0.4em\relax San Francisco, CA, USA:
  Morgan Kaufmann Publishers Inc., 2003, p. 1542–1543.

\bibitem{10.5555/3000001.3000101}
A.~Guerri and M.~Milano, ``Learning techniques for automatic algorithm
  portfolio selection,'' in \emph{Proceedings of the 16th European Conference
  on Artificial Intelligence}, ser. ECAI'04.\hskip 1em plus 0.5em minus
  0.4em\relax NLD: IOS Press, 2004, p. 475–479.

\bibitem{10.1145/1538902.1538906}
\BIBentryALTinterwordspacing
K.~Leyton-Brown, E.~Nudelman, and Y.~Shoham, ``Empirical hardness models:
  Methodology and a case study on combinatorial auctions,'' \emph{J. ACM},
  vol.~56, no.~4, jul 2009. [Online]. Available:
  \url{https://doi.org/10.1145/1538902.1538906}
\BIBentrySTDinterwordspacing

\bibitem{5365884}
Y.~Malitsky and M.~Sellmann, ``Stochastic offline programming,'' in \emph{2009
  21st IEEE International Conference on Tools with Artificial Intelligence},
  2009, pp. 784--791.

\bibitem{10.5555/2898607.2898641}
L.~Xu, H.~H. Hoos, and K.~Leyton-Brown, ``Hydra: Automatically configuring
  algorithms for portfolio-based selection,'' in \emph{Proceedings of the
  Twenty-Fourth AAAI Conference on Artificial Intelligence}, ser.
  AAAI'10.\hskip 1em plus 0.5em minus 0.4em\relax AAAI Press, 2010, p.
  210–216.

\bibitem{10.5555/1860967.1861114}
S.~Kadioglu, Y.~Malitsky, M.~Sellmann, and K.~Tierney, ``Isac
  --instance-specific algorithm configuration,'' in \emph{Proceedings of the
  2010 Conference on ECAI 2010: 19th European Conference on Artificial
  Intelligence}.\hskip 1em plus 0.5em minus 0.4em\relax NLD: IOS Press, 2010,
  p. 751–756.

\bibitem{https://doi.org/10.48550/arxiv.1906.01827}
\BIBentryALTinterwordspacing
B.~Mirzasoleiman, J.~Bilmes, and J.~Leskovec, ``Coresets for data-efficient
  training of machine learning models,'' 2019. [Online]. Available:
  \url{https://arxiv.org/abs/1906.01827}
\BIBentrySTDinterwordspacing

\bibitem{yoon2022online}
\BIBentryALTinterwordspacing
J.~Yoon, D.~Madaan, E.~Yang, and S.~J. Hwang, ``Online coreset selection for
  rehearsal-based continual learning,'' in \emph{International Conference on
  Learning Representations}, 2022. [Online]. Available:
  \url{https://openreview.net/forum?id=f9D-5WNG4Nv}
\BIBentrySTDinterwordspacing

\bibitem{https://doi.org/10.48550/arxiv.1702.08248}
\BIBentryALTinterwordspacing
O.~Bachem, M.~Lucic, and A.~Krause, ``Scalable k-means clustering via
  lightweight coresets,'' 2017. [Online]. Available:
  \url{https://arxiv.org/abs/1702.08248}
\BIBentrySTDinterwordspacing

\bibitem{https://doi.org/10.48550/arxiv.2112.09318}
\BIBentryALTinterwordspacing
B.~Wronski, ``Procedural kernel networks,'' 2021. [Online]. Available:
  \url{https://arxiv.org/abs/2112.09318}
\BIBentrySTDinterwordspacing

\bibitem{Huangfu2018}
\BIBentryALTinterwordspacing
Q.~Huangfu and J.~A.~J. Hall, ``Parallelizing the dual revised simplex
  method,'' \emph{Mathematical Programming Computation}, vol.~10, no.~1, pp.
  119--142, Mar 2018. [Online]. Available:
  \url{https://doi.org/10.1007/s12532-017-0130-5}
\BIBentrySTDinterwordspacing

\bibitem{10.1145/1772690.1772862}
\BIBentryALTinterwordspacing
D.~Sculley, ``Web-scale k-means clustering,'' in \emph{Proceedings of the 19th
  International Conference on World Wide Web}, ser. WWW '10.\hskip 1em plus
  0.5em minus 0.4em\relax New York, NY, USA: Association for Computing
  Machinery, 2010, p. 1177–1178. [Online]. Available:
  \url{https://doi.org/10.1145/1772690.1772862}
\BIBentrySTDinterwordspacing

\bibitem{46180}
\BIBentryALTinterwordspacing
D.~Golovin, B.~Solnik, S.~Moitra, G.~Kochanski, J.~E. Karro, and D.~Sculley,
  Eds., \emph{Google Vizier: A Service for Black-Box Optimization}, 2017.
  [Online]. Available:
  \url{http://www.kdd.org/kdd2017/papers/view/google-vizier-a-service-for-black-box-optimization}
\BIBentrySTDinterwordspacing

\bibitem{https://doi.org/10.48550/arxiv.1807.02811}
\BIBentryALTinterwordspacing
P.~I. Frazier, ``A tutorial on bayesian optimization,'' 2018. [Online].
  Available: \url{https://arxiv.org/abs/1807.02811}
\BIBentrySTDinterwordspacing

\bibitem{nevergrad}
J.~Rapin and O.~Teytaud, ``{Nevergrad - A gradient-free optimization
  platform},'' \url{https://GitHub.com/FacebookResearch/Nevergrad}, 2018.

\bibitem{doi:10.1126/science.aal5054}
\BIBentryALTinterwordspacing
J.~Zhang, P.~Fiers, K.~A. Witte, R.~W. Jackson, K.~L. Poggensee, C.~G. Atkeson,
  and S.~H. Collins, ``Human-in-the-loop optimization of exoskeleton assistance
  during walking,'' \emph{Science}, vol. 356, no. 6344, pp. 1280--1284, 2017.
  [Online]. Available:
  \url{https://www.science.org/doi/abs/10.1126/science.aal5054}
\BIBentrySTDinterwordspacing

\bibitem{10.2307/2332286}
\BIBentryALTinterwordspacing
W.~R. Thompson, ``On the likelihood that one unknown probability exceeds
  another in view of the evidence of two samples,'' \emph{Biometrika}, vol.~25,
  no. 3/4, pp. 285--294, 1933. [Online]. Available:
  \url{http://www.jstor.org/stable/2332286}
\BIBentrySTDinterwordspacing

\bibitem{https://doi.org/10.48550/arxiv.1707.02038}
\BIBentryALTinterwordspacing
D.~Russo, B.~Van~Roy, A.~Kazerouni, I.~Osband, and Z.~Wen, ``A tutorial on
  thompson sampling,'' 2017. [Online]. Available:
  \url{https://arxiv.org/abs/1707.02038}
\BIBentrySTDinterwordspacing

\bibitem{keselman2022fuzzy}
L.~Keselman and M.~Hebert, ``Approximate differentiable rendering with
  algebraic surfaces,'' in \emph{European Conference on Computer Vision
  (ECCV)}, 2022.

\bibitem{Lavalle98rapidly-exploringrandom}
S.~M. Lavalle, ``Rapidly-exploring random trees: A new tool for path
  planning,'' Tech. Rep., 1998.

\bibitem{Lv2018}
\BIBentryALTinterwordspacing
X.~Lv, Y.~Wang, J.~Deng, G.~Zhang, and L.~Zhang, ``Improved particle swarm
  optimization algorithm based on last-eliminated principle and enhanced
  information sharing,'' \emph{Computational Intelligence and Neuroscience},
  vol. 2018, p. 5025672, Dec 2018. [Online]. Available:
  \url{https://doi.org/10.1155/2018/5025672}
\BIBentrySTDinterwordspacing

\bibitem{4359315}
H.~Hirschmuller, ``Stereo processing by semiglobal matching and mutual
  information,'' \emph{IEEE Transactions on Pattern Analysis and Machine
  Intelligence}, vol.~30, no.~2, pp. 328--341, 2008.

\bibitem{opencv_library}
G.~Bradski, ``{The OpenCV Library},'' \emph{Dr. Dobb's Journal of Software
  Tools}, 2000.

\bibitem{https://doi.org/10.48550/arxiv.1602.07360}
\BIBentryALTinterwordspacing
F.~N. Iandola, S.~Han, M.~W. Moskewicz, K.~Ashraf, W.~J. Dally, and K.~Keutzer,
  ``Squeezenet: Alexnet-level accuracy with 50x fewer parameters and <0.5mb
  model size,'' 2016. [Online]. Available:
  \url{https://arxiv.org/abs/1602.07360}
\BIBentrySTDinterwordspacing

\bibitem{NEURIPS2019_9015}
A.~Paszke, S.~Gross, F.~Massa, A.~Lerer, J.~Bradbury, G.~Chanan, T.~Killeen,
  Z.~Lin, N.~Gimelshein, L.~Antiga, A.~Desmaison, A.~Kopf, E.~Yang, Z.~DeVito,
  M.~Raison, A.~Tejani, S.~Chilamkurthy, B.~Steiner, L.~Fang, J.~Bai, and
  S.~Chintala, ``Pytorch: An imperative style, high-performance deep learning
  library,'' in \emph{Advances in Neural Information Processing Systems 32},
  H.~Wallach, H.~Larochelle, A.~Beygelzimer, F.~d\textquotesingle
  Alch\'{e}-Buc, E.~Fox, and R.~Garnett, Eds.\hskip 1em plus 0.5em minus
  0.4em\relax Curran Associates, Inc., 2019, pp. 8024--8035.

\bibitem{6942976}
J.~D. Gammell, S.~S. Srinivasa, and T.~D. Barfoot, ``Informed rrt*: Optimal
  sampling-based path planning focused via direct sampling of an admissible
  ellipsoidal heuristic,'' in \emph{2014 IEEE/RSJ International Conference on
  Intelligent Robots and Systems}, 2014, pp. 2997--3004.

\bibitem{lai2021SbpEnv}
\BIBentryALTinterwordspacing
T.~Lai, ``sbp-env: A python package for sampling-based motion planner and
  samplers,'' \emph{Journal of Open Source Software}, vol.~6, no.~66, p. 3782,
  2021. [Online]. Available: \url{https://doi.org/10.21105/joss.03782}
\BIBentrySTDinterwordspacing

\bibitem{lai_rrdt2019}
T.~Lai, F.~Ramos, and G.~Francis, ``Balancing global exploration and
  local-connectivity exploitation with rapidly-exploring random
  disjointed-trees,'' in \emph{Proceedings of The International Conference on
  Robotics and Automation (ICRA)}.\hskip 1em plus 0.5em minus 0.4em\relax IEEE,
  2019.

\bibitem{schubert2018vidataset}
D.~Schubert, T.~Goll, N.~Demmel, V.~Usenko, J.~Stueckler, and D.~Cremers, ``The
  tum vi benchmark for evaluating visual-inertial odometry,'' in
  \emph{International Conference on Intelligent Robots and Systems (IROS)},
  October 2018.

\bibitem{stumberg22dmvio}
L.~von Stumberg and D.~Cremers, ``{DM-VIO}: Delayed marginalization
  visual-inertial odometry,'' \emph{International Conference on Robotics and
  Automation ({ICRA})}, vol.~7, no.~2, pp. 1408--1415, 2022.

\bibitem{grupp2017evo}
M.~Grupp, ``evo: Python package for the evaluation of odometry and slam.''
  \url{https://github.com/MichaelGrupp/evo}, 2017.

\bibitem{menard:hal-01475078}
\BIBentryALTinterwordspacing
P.~M{\'e}nard and A.~Garivier, ``{A minimax and asymptotically optimal
  algorithm for stochastic bandits},'' \emph{{Algorithmic Learning Theory}},
  vol.~76, 2017. [Online]. Available:
  \url{https://hal.archives-ouvertes.fr/hal-01475078}
\BIBentrySTDinterwordspacing

\end{thebibliography}

\end{document}